\documentclass[a4paper,12pt]{article}
\usepackage{graphicx} % Required for inserting images
\usepackage[T1]{fontenc}
\usepackage{lmodern}
\usepackage{parskip}
\usepackage{latexsym,amsmath,amssymb}
\usepackage{comment}
\usepackage{times}
\usepackage{zi4}
\usepackage[margin=1in]{geometry}
\usepackage{subfigure}
\usepackage{hyperref}
\usepackage{float}
\usepackage{enumerate}
\usepackage{bm,bbm}
\usepackage{listings}
\usepackage{ragged2e}

\usepackage{multirow,booktabs,setspace,float,threeparttable,tabularx}
\usepackage{makecell}
\newcolumntype{L}{>{\RaggedRight\hangafter=1\hangindent=0em}X}
\newcolumntype{C}{>{\centering\arraybackslash}X}
\usepackage{tikz}

\usepackage[ruled,vlined]{algorithm2e}

\usepackage{etoolbox}
\AtBeginEnvironment{lstlisting}{\singlespacing}

\usepackage{caption}
\usepackage{natbib}
\title{Federated Item Response Models: A Gradient-driven Privacy-preserving Framework for Distributed Psychometric Estimation}
\author{
  Biying Zhou\textsuperscript{1,**},
  Nanyu Luo\textsuperscript{1,**},
  Feng Ji\textsuperscript{1}\thanks{
    {FJ was supported by the University of Toronto Connaught Fund (FJ, grant number 520245) and Social Sciences and Humanities Research Council of Canada (FJ, grant number 518516; grant number 519992). \newline
    \indent{\textsuperscript{**}Biying Zhou and Nanyu Luo contributed equally to this work and share first authorship.}\newline
    \indent{\it Address for correspondence}: Feng Ji,
    University of Toronto, 252 Bloor St W, Toronto, ON M5S 1V6, Canada.  {\sf
    E-mail: f.ji@utoronto.ca}}
  }\\ 
  \textsuperscript{1} University of Toronto
}

\date{March 2026}

\begin{document}
\maketitle

\begin{abstract}
Item Response Theory (IRT) models are widely used to estimate respondents' latent abilities and calibrate item difficulty. Traditional IRT estimation typically requires centralizing all raw responses, raising privacy and governance concerns. We introduce Federated Item Response Theory (FedIRT), a framework that enables distributed calibration of standard IRT models without transferring individual-level data, thereby preserving confidentiality while retaining statistical efficiency. 

To provide formal protection, we further develop FedIRT-DP, a user-level differentially private extension. Each site computes per-student gradients, clips them to a fixed norm, and shares only masked sums; the server adds calibrated Gaussian noise and performs MAP updates. This yields an auditable \((\varepsilon,\delta)\) guarantee at the student level and a single, tunable privacy-utility trade-off via the clipping bound and noise scale. The same mechanism improves robustness to extreme response rows (e.g., all-zeros/ones).

Across simulations, FedIRT matches the accuracy of centralized estimators from popular \texttt{R} packages while avoiding data pooling; FedIRT-DP achieves comparable accuracy under stronger privacy and exhibits superior robustness to contamination. An empirical study on a real exam dataset demonstrates practical viability and consistent item and site-effect estimates. To facilitate adoption, we release an open-source \texttt{R} package, \texttt{FedIRT}, implementing the two-parameter logistic (2PL) and partial credit models (PCM) with federated and differentially private training.

\end{abstract}

\doublespacing
\setlength{\abovedisplayskip}{6pt}
\setlength{\belowdisplayskip}{6pt}
\setlength{\abovedisplayshortskip}{4pt}
\setlength{\belowdisplayshortskip}{4pt}
\setlength\jot{4pt}%
\newpage

\section{Introduction}
Preserving student privacy has become a critical priority in educational assessment. With increasing digitalization, safeguarding sensitive information such as students' test responses has gained global attention through legal measures and institutional policies. Legal frameworks across the globe now enforce strict standards for handling student data. In Europe, the General Data Protection Regulation \citep[GDPR,][]{regulation2016regulation} requires explicit consent for processing minors' data and limits how schools collect, use, and store student records. Similarly, the Family Educational Rights and Privacy Act (FERPA) in the United States \citep{rights2014family} prohibits sharing personally identifiable student information without consent, granting students full control over their records after age 18. These regulations are further supported by voluntary initiatives like the Student Privacy Pledge, reflecting a worldwide effort to strengthen data security in education.

Item Response Theory (IRT) remains a cornerstone of test development and validation across education, psychology, and health science because of its flexibility and interpretability \citep[e.g.,][]{embretson2013item}. Yet, privacy considerations have rarely been explored or discussed within the IRT literature as well as in general psychometrics literature. Notable exceptions include \citet{dwork2006differential}, who introduced Differential Privacy (DP) to protect individuals in statistical databases; \citet{lemons2014predictive}, who examined data swapping and its influence on differential item functioning (DIF); and \citet{nguyen2023optimal}, who proposed a discrete Gaussian mechanism that balances privacy and accuracy. Still, these studies stop short of offering a unified framework applicable to general IRT models and many of these studies focus solely on the conceptual or technical aspects therefore lacking practical utility.

Federated learning provides a promising framework for privacy-preserving data analysis which has great potential for psychometric analysis with additional privacy. First introduced by Google in 2016 \citep{konevcny2016federated}, this method trains models across decentralized datasets without transferring raw data. Participants receive model parameters from a central server, compute updates locally, and share only these updates. By keeping data on-site, federated learning minimizes privacy risks \citep{kairouz2021advances} and communication costs \citep{yang2019federated}, making it ideal for sensitive contexts like data in education.

Furthermore, federated learning can be combined with differential privacy to provide a formal, auditable guarantee that limits what can be learned about any one student from the analysis, regardless of any side information an adversary might hold. To be specific, \emph{user-level} DP, which can be adopted to protect a student's \emph{entire} response vector rather than a single record, has two properties that are especially valuable in operational assessment: post-processing immunity (downstream computations cannot weaken the guarantee) and composability (privacy loss accumulates in a controlled way across rounds of estimation)~\citep{dwork2014algorithmic}. These features allow institutions to publish a clear $(\varepsilon,\delta)$ privacy budget and to trade off accuracy and privacy in a transparent manner that aligns with GDPR/FERPA compliance.

Building on these advances, we introduce \emph{Federated IRT}, a framework that integrates federated learning with marginal maximum likelihood estimation for distributed calibration of IRT models. Methodologically, the approach instantiates distributed maximum-likelihood estimation, but its contribution is to formalize the link between federated learning and IRT through a common gradient-aggregation mechanism. Since this connection is very first exploration of privacy-preserving procedures in IRT estimation through federated learning, our contributions are twofold. First, Federated IRT addresses the growing constraints on centralized data collection by enabling multi-institution estimation without transferring students' raw responses. Second, the framework accommodates general IRT formulations, permitting the simultaneous and precise estimation of item parameters and school-level effects from summary statistics alone. This capability holds particular promise for education, psychology, and the social science, where data often reside in separate institutions and require analysis of item characteristics (e.g., difficulty, discrimination) and group differences without centralizing sensitive records.

Beyond federated estimation, another contribution is that we develop \emph{FedIRT-DP}, an novel, end-to-end user-level differential privacy extension for IRT. Each school computes \emph{per-student} gradients. This design improves robustness to extreme response rows through the same clipping that enforces DP. FedIRT-DP delivers estimates comparable to federated baselines under regular conditions and degrades gracefully under contamination, an important practical benefit when responses include guessing, copying, or other atypical patterns.

The remainder of the paper is organized as follows. Section~2 reviews the federated learning paradigm, its core algorithms, and recent statistical applications. Section~3 details the Federated IRT methodology, including school-level extensions and the implementation of marginal maximum likelihood within the federated setting, and presents the FedIRT-DP privacy accounting. Sections~4 and~5 present simulation and empirical studies that demonstrate the efficacy of the proposed approach. We close with a discussion of implications and directions for future research.

\section{Federated Learning Framework}
Federated learning is a methodological framework that enables collaborative model training across multiple clients or institutions by keeping raw data locally and sharing only model updates.  In what follows, we first describe two foundational algorithms, Federated Averaging and Federated Stochastic Gradient Descent, that form the core of most federated learning systems. We then review recent efforts to apply federated principles to classical statistical models, and finally introduce our own Federated Item Response Theory framework for privacy-preserving psychometric estimation.
\subsection{Federated Averaging (FedAvg)}
Federated Averaging \citep[FedAvg,][]{mcmahan2017communication, li2019convergence} is a fundamental algorithm in federated learning that enables multiple clients to collaboratively train a shared model by averaging locally computed updates, thus preserving data privacy and reducing central storage requirements. Formally, FedAvg aims to minimize the global objective function:
\begin{equation}\label{eq:global_objective}
    F(w) = \sum_{k=1}^{K} \frac{N_k}{N}\,F_k(w),
\end{equation}
where $F_k(w)$ is the loss on client $k$'s local dataset of size $N_k$, and $N = \sum_{k=1}^K N_k$. 

During each round $t$, the server samples a fraction $C\in(0,1]$ of clients, broadcasts the current parameters $w^{(t)}$, and each selected client $k$ performs $E$ steps of local stochastic gradient descent on its own data:
\begin{equation}
    w_k^{(t,0)} = w^{(t)},\qquad
    w_k^{(t,e+1)} = w_k^{(t,e)} - \eta g_k\bigl(w_k^{(t,e)}\bigr),\quad e=0,\ldots,E-1,
\end{equation}
with mini-batch gradients of the loss function $g_k(\cdot)$ and learning rate $\eta$. After the local updates, client $k$ returns $w_k^{(t+1)}=w_k^{(t,E)}$. Let $\mathcal{S}^{(t)}$ denote the participating clients and $N_{\mathcal{S}^{(t)}}=\sum_{j\in\mathcal{S}^{(t)}} N_j$. The server forms the next global iterate by a size-weighted average:
\begin{equation}\label{eq:fedavg_param}
w^{(t+1)}=\sum_{k\in\mathcal{S}^{(t)}} \frac{N_k}{N_{\mathcal{S}^{(t)}}} w_k^{(t+1)}.
\end{equation}

The hyperparameters $C$ (client fraction) and $E$ (local epochs) allow practitioners to balance convergence speed against communication cost in heterogeneous environments. FedAvg achieves a favorable balance between optimization efficiency and privacy protection: only model parameters are exchanged (not raw data), and the aggregation step requires only a simple, easily computed weighted average.

\subsection{Federated Stochastic Gradient Descent (FedSGD)}
Federated Stochastic Gradient Descent \citep[FedSGD,][]{mcmahan2017communication} is a simpler variant in which each selected client computes exactly one gradient on its entire local dataset and immediately shares that gradient, rather than performing multiple local updates. FedSGD optimizes the same global objective as Eq.~\ref{eq:global_objective} for FedAvg, but updates the model by a single aggregated gradient step per round. In iteration $t$, after broadcasting $w^{(t)}$, each client $k$ computes the mini-batch gradient $g_k^{(t)} = g_k(w^{(t)})$ and sends $g_k^{(t)}$ to the server. The server then updates via
\begin{equation}
    w^{(t+1)} = w^{(t)} - \eta \sum_{k\in\mathcal{S}^{(t)}} \frac{N_k}{N_{\mathcal{S}^{(t)}}} g_k^{(t)}.
\end{equation}
Because each client performs only one gradient computation, this algorithm can be slow to converge due to systems heterogeneity and non-identified data distribution among clients. One inherent issue is its communication inefficiency since it requires frequent transmission of model updates between sites and the server, and trades off more required rounds for faster per-round execution. Such inefficiency becomes a significant factor considering the limited network bandwidth and high heterogeneity often associated with federated learning settings \citep{li2020federated}, but FedSGD retains conceptual alignment with classic Stochastic Gradient Descent algorithm and serves as a useful baseline. 

\subsection{Federated Learning for Statistical Models}
Federated learning has increasingly been applied in the field of applied statistics \cite[e.g.,][]{brisimi2018federated, sheller2020federated, huang2020loadaboost}. One notable application involves the development of Generalized Linear Mixed Models (GLMMs), which are widely used to analyze patient data in medical research. Several studies have explored the construction of GLMMs through federated learning frameworks to manage hospital data while maintaining privacy \citep{li2022federated, yan2023privacy, luo2022dpql}. These approaches typically use the likelihood function to formulate the loss function within the federated learning algorithm and perform Maximum Likelihood Estimation (MLE) of model parameters through iterative likelihood and gradient updates, similar in spirit to the FedAvg and FedSGD algorithms.

\citet{a15070243} introduced a broader federated learning framework for distributed parameter estimation in Generalized Linear Models (GLMs). Their work is primarily theoretical, focusing on extending the estimation process to accommodate various types of GLMs within federated environments. They proposed a novel technique that combines parameter estimates from multiple institutions using a weighted averaging method designed to minimize both bias and variance in the final model. In addition, the study analyzed the computational efficiency of the proposed framework, demonstrating its applicability to a wide range of GLM problems beyond hospital data settings.

Federated learning combined with MLE represents a growing area within the broader federated learning literature \citep{liu2020systematic}. In general, the central server collects gradients of the likelihood function from each client to update global model parameters and subsequently sends the updated parameters back to the clients. Through repeated iterations, the overall model gradually converges. Motivated by this approach, as the IRT model is a specific class of GLMs, we propose a new framework, termed \textit{Federated Item Response Theory} (FedIRT), to perform estimation in a federated learning setting. The FedIRT framework is designed to estimate item parameters and school-level effects across multiple institutions while safeguarding data privacy. This method enables the analysis of educational assessments without requiring the centralization of sensitive student information, thereby addressing privacy concerns inherent in traditional IRT modeling. A detailed comparison among FedAvg, FedSGD, and FedIRT is presented in Table~\ref{tab:alg_comparison}.
\begin{table}[!htb]
    \centering
    \caption{Comparison of key components across FedAvg, FedSGD, and FedIRT.}
    \label{tab:alg_comparison}
    \begin{tabular}{p{1.5cm} p{4.5cm} p{5cm} p{3.7cm}}
    \toprule
    \textbf{Name}  & \textbf{Objective function} & \textbf{Gradient computation}  & \textbf{Shared parameters}  \\
    \midrule
    FedAvg & $F(w) = \sum_{k = 1} ^ {K} \frac{N_k}{N} F_k(w)$ & Only $g_k(w^{(t)})$ & $w^{(t)},w_k^{(t)}$  \\
    \addlinespace[0.5em]
    FedSGD & $F(w) = \sum_{k = 1} ^ {K} \frac{N_k}{N} F_k(w)$ & $ g(w^{(t)}) = \sum_{k\in\mathcal{S}^{(t)}} \frac{N_k}{N_{\mathcal{S}^{(t)}}} g_k^{(t)}$ & $w^{(t)}, g_k^{(t)}$\\
    \addlinespace[0.5em]
    FedIRT & $\ell = \sum_{k=1}^{K}\ell_k$ & 
    $\displaystyle\frac{\partial \ell}{\partial (\alpha, \beta, s)} = \sum\limits_{k=1}^{K} \frac{\partial \ell_k}{\partial (\alpha, \beta, s)}$ & 
    $ \alpha_j$, $\beta_j$, $ s_k$, $\displaystyle\frac{\partial \ell_k}{\partial (\alpha, \beta, s)}$ \\
    \bottomrule
    \end{tabular}
    \label{compare fed-learning models}
\end{table}

\section{Federated IRT}
Federated Item Response Theory framework brings together the privacy-preserving benefits of federated learning and the measurement framework of item response theory. This section elaborates on the incorporation of school-level effects into IRT models and the implementation of MLE within a federated learning context. To improve the privacy protection of the proposed framework, we also extend FedIRT to incorporate differential privacy, resulting in FedIRT-DP.
\subsection{Item Response Theory Models with School-Level Effects}
Item Response Theory models provide a robust framework for understanding the relationship between latent traits and observed test responses. In the standard \textit{two-parameter logistic} (2PL) model, as introduced by \citet{bock1981marginal}, item characteristics including discrimination and difficulty are treated as fixed effects, while individual abilities are modeled as random effects. However, when data are collected from multiple schools, it is unrealistic to assume that all schools share the same average ability level. In our approach, we posit that a student’s performance is influenced not only by their individual ability but also by the unique effect of the school they attend. Consequently, we extend the 2PL model by incorporating a fixed school effect.

Assume that there are $K$ schools, and within each school $k$ there are $N_k$ students, so that the total number of students across all schools is given by $N = \sum_{k=1}^K N_k$. Each student completes an exam consisting of $J$ items. Let the response for student $i$ in school $k$ to item $j$ be denoted by $x_{ijk} \in \{0,1\}$, where $x_{ijk}=1$ indicates a correct answer, and $x_{ijk}=0$ indicates an incorrect answer. The parameters associated with each item include a discrimination parameter $\alpha_j$ and a difficulty parameter $\beta_j$, both assumed invariant across schools. Each student is assumed to possess an ability $\theta_{ik}$, and each school has its own effect denoted by $s_k$\footnote{This fixed-effect term has appeared in some explantory IRT models \citep[e.g.,][]{de2004explanatory} but the purpose here is to help interpret institutional differences.}, which can also be refereed to as school level characteristics that are associated with students' ability or school effect. For modeling purposes, the individual abilities $\theta_{ik}$ are assumed to follow a standard normal distribution.

The probability that a student with ability $\theta_{ik}$ in school $ k $ correctly responds to item $ j $ is modeled using the logistic function:
\begin{equation}\label{eq:2PL}
    P_j(x_{ijk} = 1 \mid \theta_{ik}) = \sigma \bigl(\alpha_j(\theta_{ik} + s_k - \beta_j)\bigr) = \frac{\exp\bigl(\alpha_j(\theta_{ik} + s_k - \beta_j)\bigr)}{1 + \exp\bigl(\alpha_j(\theta_{ik} + s_k - \beta_j)\bigr)}.
\end{equation}
Notice that the school effect $ s_k $ is additive with the student's ability $\theta_{ik}$ in the logit scale, thereby directly contributing the probability of a correct response. Moreover, if one constrains all discrimination parameters to one (i.e., $\alpha_j = 1$ for all $ j $), this formulation reduces to the one-parameter logistic (1PL) model.

While the 2PL model is appropriate for dichotomously scored items, many educational and psychological assessments involve items that are scored across multiple categories. In such cases, the \textit{Partial Credit Model} (PCM), as described by \citet{masters1982rasch}, is more suitable. In order to account for the school effect within this context, we modify the model by incorporating the fixed school effect $ s_k $ into the latent trait component. This modification ensures that the school-level influences are reflected in the probability of observing a particular score on each item. Given the maximum score possible for item $ j $ is $ C_j-1 $, the PCM models the probability that an individual with ability $\theta_{ik}$ attains a score of $ z \in \{0,\cdots,C_j-1\}$ on item $ j $ as follows:
\begin{equation}
    \pi_{ijkz}:=P_j(x_{ijk} = z \mid \theta_{ik}) = \frac{\exp\bigl(\sum_{h=0}^{z} \alpha_j(\theta_{ik} + s_k - \beta_{jh})\bigr)}{\sum_{c=0}^{C_j-1} \exp\bigl(\sum_{h=0}^{c} \alpha_j(\theta_{ik} + s_k - \beta_{jh})\bigr)}.
\end{equation}
For $ z=0 $, $\sum^0_{h=0}\alpha_j(\theta_{ik} + s_k - \beta_{jh})=0$. If $C_j=2$, then PCM reduces to the 2PL model.

\subsection{Marginal Maximum Likelihood Estimation for FedIRT}
\textit{Marginal Maximum Likelihood Estimation} \citep[MMLE,][]{bock1981marginal} has become a prevalent framework for recovering both item and latent-trait parameters in IRT. Under this framework, we estimate discrimination parameters $\bm{\alpha}=(\alpha_1,\dots,\alpha_J)$, difficulty parameters $\bm{\beta}=(\beta_1,\dots,\beta_J)$, and school-level effects $\bm{s}=(s_1,\dots,s_k)$ by maximizing the marginal likelihood of all observed responses.

For student $i$ in school $k$, the response to item $j$, denoted $x_{ijk}$, follows a multinomial distribution with $C_j$ categories, a single trial, and cell probabilities $\{\pi_{ijkz}\}_{z=0}^{C_j-1}$:
\begin{equation}
    p_j(x_{ijk} \mid \theta_{ik}) = \prod_{z=0}^{C_j-1}\pi_{ijkz}^{\mathbbm{1}(x_{ijk}=z)},
\end{equation}
where the indicator function $\mathbbm{1}(x_{ijk}=z)=1$ if $x_{ijk}=z$ otherwise $\mathbbm{1}(x_{ijk}=z)=0$ for $z \in \{0,\dots,C_j-1\}$. Denoting the full response vector for student $i$ by $\bm{x}_{ik}=(x_{i1k},\dots,x_{iJk})^T$, and invoking conditional independence given ability $\theta_{ik}$, the conditional distribution of $\bm{x}_{ik}$ is
\begin{equation}
    p(\bm{x}_{ik} \mid \theta_{ik})=\prod_{j=1}^{J}p_j(x_{ijk} \mid \theta_{ik}).
\end{equation}

Assuming abilities are random draws from a standard normal distribution, $\theta_{ik}\sim\mathcal{N}(0,1)$ with density $g(\theta)$, the marginal probability of $\bm{x}_{ik}$ becomes
\begin{equation}
    p(\bm{x}_{ik})=\int_{-\infty}^{\infty}p(\bm{x}_{ik} \mid \theta_{ik})g(\theta_{ik})d\theta_{ik}=\int_{-\infty}^{\infty}\prod_{j=1}^{J}p_j(x_{ijk} \mid \theta_{ik})g(\theta_{ik})d\theta_{ik}.
\end{equation}

Let $X_k$ be an $N_k\times J$ matrix of independent responses whose $i^\text{th}$ row is $\bm{x}_{ik}^T$. For each school $k$, the observed likelihood contribution of its $N_k$ students is
\begin{equation}
    \mathcal{L}_k(\bm{\alpha},\bm{\beta},s_k\mid X_k) = \prod_{i=1}^{N_k}p(\bm{x}_{ik}) = \prod_{i=1}^{N_k}\left[\int_{-\infty}^{\infty}\prod_{j=1}^{J}p_j(x_{ijk} \mid \theta_{ik})g(\theta_{ik})d\theta_{ik}\right].
\end{equation}
Given $\bm{X}$ collecting all $X_k$'s, aggregating across all $k$ schools yields the overall likelihood: 
\begin{equation}
    \mathcal{L}(\bm{\alpha},\bm{\beta},\bm{s}\mid \bm{X}) = \prod_{k=1}^{K}\mathcal{L}_k(\bm{\alpha},\bm{\beta},s_k\mid X_k).
\end{equation}
Maximizing $\mathcal{L}(\bm{\alpha},\bm{\beta},\bm{s}\mid X_k)$ directly is difficult because we must evaluate the $N$ integrals. Using Gaussian-Hermite quadrature \citep{bojanov1986generalized} to obtain an approximated value is a classical way to calculate integral numerically. By using he Gaussian-Hermite quadrature, we can assume that all students have $\theta \in [-\theta_0,\theta_0]$. We divide the population into $q$ equally-spaced levels $n=1,\dots,q$ \citep{chalmers2012mirt} and assume that in each level, people share the same ability. We define $V(n)$ as the ability value of level $n$, and $A(n)$ is the weight of level $n$. Then the marginal distribution of $\bm{x}_{ik}$ can be approximated by:
\begin{equation}\label{eq:approximate_marginal}
    \begin{aligned}
        p(\bm{x}_{ik})&=\int_{-\infty}^{\infty}p(\bm{x}_{ik} \mid \theta_{ik})g(\theta_{ik})d\theta_{ik} \approx \sum_{n=1}^{q} p(\bm{x}_{ik} \mid \theta_{ik}=V
    (n))A(n) := \tilde{p}(\bm{x}_{ik}),\\
        V(n) &= -\theta_0 + (n - 1) \frac{\theta_0 - (-\theta_0)}{q - 1} = -\theta_0 + \frac{2\theta_0(n-1)}{q-1},\\
    A(n) &=  \int_{V(n) - \frac{\theta_0 - (-\theta_0)}{2(q - 1)}} ^{V(n) + \frac{\theta_0 - (-\theta_0)}{2(q - 1)}} g(\theta) d\theta = \int_{V(n) - \frac{\theta_0}{q - 1}} ^{V(n) + \frac{\theta_0 }{q - 1}} g(\theta) d\theta. 
    \end{aligned}
\end{equation}

Consequently, the overall log-likelihood can be approximated by:
\begin{equation}
    \begin{aligned}
    \ell(\bm{\alpha},\bm{\beta},\bm{s}\mid \bm{X}) &\approx \sum_{k=1}^{K}\sum_{i=1}^{N_k}\log \tilde{p}(\bm{x}_{ik})
    \end{aligned}
\end{equation}

Let $u_j$ denote a generic item parameter for item $j$.  To achieve the maximum likelihood estimates of all parameters, we can use the \textit{Expectation-Maximization} (EM) algorithm introduced by \citet{bock1981marginal} to solve: for each $j=1,\dots,J$ and $k=1,\dots,K$,
\begin{equation}
    \begin{aligned}
        \frac{\partial\ell(\bm{\alpha},\bm{\beta},\bm{s}\mid \bm{X})}{\partial u_j}&=\sum_{k=1}^{K} \frac{\partial\ell_k(\bm{\alpha},\bm{\beta},s_k\mid X_k)}{\partial u_j}=\sum_{k=1}^{K} \sum_{i=1}^{N_k} \frac{\partial}{\partial u_j}\log \tilde{p}(\bm{x}_{ik})=0,\\
        \frac{\partial\ell(\bm{\alpha},\bm{\beta},\bm{s}\mid \bm{X})}{\partial s_k}&=\frac{\partial\ell_k(\bm{\alpha},\bm{\beta},s_k\mid X_k)}{\partial s_k}=\sum_{i=1}^{N_k} \frac{\partial}{\partial s_k}\log \tilde{p}(\bm{x}_{ik})=0,
    \end{aligned}
\end{equation}

We consider the 2PL simple model case and denote $\pi_{ijk}(n):=P_j(x_{ijk} = 1 \mid \theta_{ik}=V(n))$. Although indexed by $i$, $\pi_{ijk}(n)$ is constant across $i=1,\dots,N_k$, since each examinee in school $k$ is assumed to have ability $\theta_{ik}=V(n)$ at quadrature node $n$. The logarithm of $\tilde{p}(\bm{x}_{ik})$ can be written as
\begin{equation}\label{eq:loglikelihood_school}
    \begin{aligned}
        \log \tilde{p}(\bm{x}_{ik}) &\approx \log\left[\sum_{n=1}^{q} p(\bm{x}_{ik} \mid \theta_{ik}=V(n))A(n)\right]\\
        &=\log\left\{\sum_{n=1}^{q} \bigl[\prod_{j=1}^{J}\pi_{ijk}(n)^{x_{ijk}}\bigl(1-\pi_{ijk}(n)\bigr)^{1-x_{ijk}}\bigr]A(n)\right\}.
    \end{aligned}
\end{equation}
Differentiating it with respect to $u_j$ gives:
\begin{equation}\label{eq:partial_likelihood}
    \begin{aligned}
    \frac{\partial}{\partial u_j}\log \tilde{p}(\bm{x}_{ik})
    \approx&\frac{1}{\tilde{p}(\bm{x}_{ik})}\left\{\sum_{n=1}^{q} \bigl[\frac{x_{ijk}}{\pi_{ijk}(n)}-\frac{1-x_{ijk}}{1-\pi_{ijk}(n)}\bigr]p(\bm{x}_{ik} \mid \theta_{ik}=V(n))\frac{\partial \pi_{ijk}(n)}{\partial u_j}A(n)\right\}\\
    =&\frac{1}{\tilde{p}(\bm{x}_{ik})}\left\{\sum_{n=1}^{q} \frac{x_{ijk}-\pi_{ijk}(n)}{\pi_{ijk}(n)\bigl(1-\pi_{ijk}(n)\bigr)}p(\bm{x}_{ik} \mid \theta_{ik}=V(n))\frac{\partial \pi_{ijk}(n)}{\partial u_j}A(n)\right\}.
    \end{aligned}
\end{equation}
Denoting the bin posterior probability of ability level $n$ for student $i$ in school $k$ by
\begin{equation}\label{eq:bin_posterior}
    p_{ik}(n) = \frac{p(\bm{x}_{ik} \mid \theta_{ik}=V(n))A(n)}{\tilde{p}(\bm{x}_{ik})}.
\end{equation}
% The ``expected sample size'' at level $n$ given $X_{k}$ is defined as:
% \begin{equation}\label{eq:expected_sample_size}
%     \hat{N}_{k}(n) = \sum_{i=1}^{N_k}\frac{p(\bm{x}_{ik} \mid \theta_{ik}=V(n))A(n)}{\tilde{p}(\bm{x}_{ik})}.
% \end{equation}
% The ``expected frequency'' of correct response to item $j$ at level $n$ given $\{x_{ijk}|i=1,\dots,N_k\}$ is defined as:
% \begin{equation}\label{eq:expected_frequency}
%     \hat{r}_{jk}(n) = \sum_{i=1}^{N_k}\frac{x_{ijk}p(\bm{x}_{ik} \mid \theta_{ik}=V(n))A(n)}{\tilde{p}(\bm{x}_{ik})}.
% \end{equation}
% With Eq. \ref{eq:partial_likelihood}, Eq. \ref{eq:expected_sample_size} and Eq. \ref{eq:expected_frequency}, and 
With Eq. \ref{eq:partial_likelihood}, Eq. \ref{eq:bin_posterior}, and
\begin{equation}
    \begin{aligned}
        \frac{\partial \pi_{ijk}(n)}{\partial \alpha_j}&=\pi_{ijk}(n)\bigl(1-\pi_{ijk}(n)\bigr)(V(n) + s_k - \beta_j),\\
        \frac{\partial \pi_{ijk}(n)}{\partial \beta_j}&=-\pi_{ijk}(n)\bigl(1-\pi_{ijk}(n)\bigr)\alpha_j,\\
    \end{aligned}
\end{equation}
we can derive the per-student contribution to the gradients: for the discrimination parameter $\alpha_j$ and difficulty parameter $\beta_j$,
\begin{align}
    g_{ik}^{(\alpha_j)} := \frac{\partial}{\partial \alpha_j}\log \tilde{p}(\bm{x}_{ik})=&\frac{1}{\tilde{p}(\bm{x}_{ik})}\left\{\sum_{n=1}^{q} \bigl(x_{ijk}-\pi_{ijk}(n) \bigr)p(\bm{x}_{ik} \mid \theta_{ik}=V(n))A(n)(V(n) + s_k - \beta_j)\right\}\nonumber\\
    =&\sum_{n=1}^{q} \bigl(x_{ijk}-\pi_{ijk}(n) \bigr)(V(n) + s_k - \beta_j)p_{ik}(n),\label{eq:grad_alpha}\\
    g_{ik}^{(\beta_j)} := \frac{\partial}{\partial \beta_j}\log \tilde{p}(\bm{x}_{ik})=&\frac{1}{\tilde{p}(\bm{x}_{ik})}\left\{-\sum_{n=1}^{q} \bigl(x_{ijk}-\pi_{ijk}(n) \bigr)p(\bm{x}_{ik} \mid \theta_{ik}=V(n))A(n)\alpha_j\right\}\nonumber\\
    =&-\alpha_j\sum_{n=1}^{q} \bigl(x_{ijk}-\pi_{ijk}(n) \bigr) p_{ik}(n)\label{eq:grad_beta}.
\end{align}
% \begin{align}
%     &\frac{\partial\ell_k(\bm{\alpha},\bm{\beta},s_k\mid \bm{X})}{\partial \alpha_j}\nonumber\\
%     \approx&\sum_{i=1}^{N_k}\frac{1}{\tilde{p}(\bm{x}_{ik})}\left\{\sum_{n=1}^{q} \bigl(x_{ijk}-\pi_{ijk}(n) \bigr)p(\bm{x}_{ik} \mid \theta_{ik}=V(n))A(n)(V(n) + s_k - \beta_j)\right\}\nonumber\\
%     =&\sum_{n=1}^{q}\left\{\sum_{i=1}^{N_k}\frac{x_{ijk}p(\bm{x}_{ik} \mid \theta_{ik}=V(n))A(n)}{\tilde{p}(\bm{x}_{ik})}-\sum_{i=1}^{N_k}\frac{\pi_{ijk}(n)p(\bm{x}_{ik} \mid \theta_{ik}=V(n))A(n)}{\tilde{p}(\bm{x}_{ik})}\right\}(V(n) + s_k - \beta_j)\nonumber\\
%     =&\sum_{n=1}^{q}\left\{\hat{r}_{jk}(n)-\pi_{ijk}(n)\hat{N}_{k}(n)\right\}(V(n) + s_k - \beta_j),\\
%     \nonumber\\
%     &\frac{\partial\ell_k(\bm{\alpha},\bm{\beta},s_k\mid \bm{X})}{\partial \beta_j}\nonumber\\
%     \approx&\sum_{i=1}^{N_k}\frac{1}{\tilde{p}(\bm{x}_{ik})}\left\{-\sum_{n=1}^{q} \bigl(x_{ijk}-\pi_{ijk}(n) \bigr)p(\bm{x}_{ik} \mid \theta_{ik}=V(n))A(n)\alpha_j\right\}\nonumber\\
%     =&-\sum_{n=1}^{q}\left\{\sum_{i=1}^{N_k}\frac{x_{ijk}p(\bm{x}_{ik} \mid \theta_{ik}=V(n))A(n)}{\tilde{p}(\bm{x}_{ik})}-\sum_{i=1}^{N_k}\frac{\pi_{ijk}(n)p(\bm{x}_{ik} \mid \theta_{ik}=V(n))A(n)}{\tilde{p}(\bm{x}_{ik})}\right\}\alpha_j\nonumber\\
%     =&-\alpha_j\sum_{n=1}^{q}\left\{\hat{r}_{jk}(n)-\pi_{ijk}(n)\hat{N}_{k}(n)\right\}.
% \end{align}

An analogous derivative for the school effect $s_k$ follows, aggregating information across all items:
\begin{equation}\label{eq:grad_school_effect}
    g_{ik}^{(s_k)} := \frac{\partial}{\partial s_k}\log \tilde{p}(\bm{x}_{ik})=\sum_{j=1}^{J}\alpha_j\sum_{n=1}^{q} \bigl(x_{ijk}-\pi_{ijk}(n) \bigr) p_{ik}(n).
\end{equation}

% \begin{equation}
%     \frac{\partial\ell_k(\bm{\alpha},\bm{\beta},s_k\mid \bm{X})}{\partial s_k}\approx \sum_{j=1}^{J}\alpha_j\sum_{n=1}^{q}\left\{\hat{r}_{jk}(n)-\pi_{ijk}(n)\hat{N}_{k}(n)\right\}.
% \end{equation}
Estimating the item parameters $\bm{\alpha}$, $\bm{\beta}$ and the school effect $s_k$ by the EM algorithm is an itereative process. In the Expectation (E) step, provisional estimates of the parameters are used to compute the bin-specific posterior probabilities $p_{ik}(n)$ for each student $i$ in school $k$. These posterior probabilities are then aggregated to yield the expected gradients of the complete-data log-likelihood. In the subsequent Maximization (M) step, these gradients are utilized to optimize the log-likelihood function, typically through numerical methods such as the Broyden-Fletcher-Goldfarb-Shanno \citep[BFGS;][]{NocedalWright2006} algorithm, which updates the parameter estimates. The E and M steps alternate until convergence is attained, as indicated by negligible changes in the log-likelihood or in successive parameter updates. With these estimation results, we can also estimate students' abilities via expected a posteriori and the standard error of parameter estimates (see Appendix).

\subsection{Implementation of Federated IRT}
\begin{figure}[!htb]
    \centering
    \includegraphics[width=\linewidth]{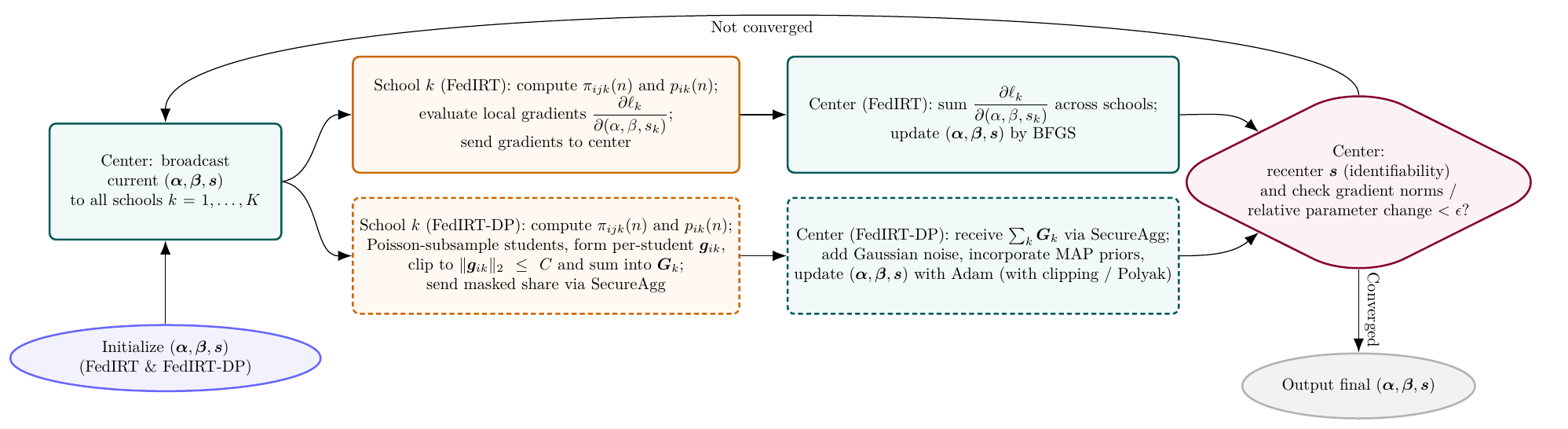}
    \caption{Flow chart for the estimation of item parameters $\bm{\alpha}$ and $\bm{\beta}$ and school effect $\bm{s}$ in the 2PL model via the FedIRT and FedIRT-DP framework.}
    \label{fig:flowchart}
\end{figure}

Building upon the marginal maximum likelihood estimation detailed above, the FedIRT framework designs a decentralized yet cohesive procedure for estimating both item parameters and school-level effects across $k$ distinct institutions. As illustrated in Fig.~\ref{fig:flowchart}, the process begins at the central server, which assigns initial values $\alpha_j = 1$ and $\beta_j = 0$ for every item $j$, and $s_k = 0$ for every school $k$. These provisional estimates are then broadcast to all participating sites.

At each school $k$, the local routine (Alg.~\ref{alg:school}) takes the response matrix $X_k$ and the incoming parameters as inputs. First, it computes the probabilities of correct responses $\pi_{ijk}(n)$ for each latent quadrature bin $n$, then derives the bin posterior weights $p_{ik}(n)$ in Eq.~\ref{eq:bin_posterior}. Using these values, the site evaluates its contribution to the overall gradients $\partial \ell_k / \partial \alpha_j$, $\partial \ell_k / \partial \beta_j$, and $\partial \ell_k / \partial s_k$. Only these summary quantities are returned to the central server, thereby preserving each school's raw data privacy.
\begin{algorithm}[!htb]
    \caption{FedIRT---Gradient computation algorithm in school $k$, $k=1,\dots,K$}
    \label{alg:school}
    \SetAlgoLined

    \SetKwInOut{Given}{Given}
    \Given{$V(n), A(n), n=1,\dots,q$}
    \KwIn{Response matrices $X_k$, item parameters $\alpha_j$, $\beta_j$, $j=1,\dots,J$, school effect $s_k$}
    \KwOut{Gradients $\frac{\partial \ell_k}{\partial \alpha_j}, \frac{\partial \ell_k}{\partial \beta_j}, j=1,\dots,J$ and $\frac{\partial \ell_k}{\partial s_k}$}
    
    Compute $\pi_{ijk}(n)$ for $i=1,\dots,N_k$, $j=1,\dots,J$\;
    Compute bin posterior probabilities $p_{ik}(n)$ for $n=1,\dots,q$\ according to Eq.~\ref{eq:bin_posterior}\;
    \For{item $j=1$ \KwTo $J$}{
    % $ \ell_k \approx \sum_{i=1}^{N_k}\log\left\{\sum_{n=1}^{q} \bigl[\prod_{j=1}^{J}\pi_{ijk}(n)^{x_{ijk}}\bigl(1-\pi_{ijk}(n)\bigr)^{1-x_{ijk}}\bigr]A(n)\right\}$\;
    $ \frac{ \partial \ell_k}  { \partial \alpha_j } = \sum_{i=1}^{N_k} g_{ik}^{(\alpha_j)} = \sum_{i=1}^{N_k}\sum_{n=1}^{q}\bigl(x_{ijk}-\pi_{ijk}(n) \bigr)(V(n) + s_k - \beta_j)p_{ik}(n)$\;
    $ \frac{ \partial \ell_k}  { \partial \beta_j } = \sum_{i=1}^{N_k} g_{ik}^{(\beta_j)}= -\sum_{i=1}^{N_k} \alpha_j\sum_{n=1}^{q}\bigl(x_{ijk}-\pi_{ijk}(n) \bigr) p_{ik}(n)$\;
    }
    $\frac{ \partial \ell_k}  { \partial s_k} =\sum_{i=1}^{N_k} g_{ik}^{(s_k)} = \sum_{i=1}^{N_k}\sum_{j=1}^{J}\alpha_j\sum_{n=1}^{q}\bigl(x_{ijk}-\pi_{ijk}(n) \bigr) p_{ik}(n)$\;
    
    Send $\frac{ \partial \ell_k}  { \partial \alpha_j }$, $\frac{ \partial \ell_k}  { \partial \beta_j }$, and $\frac{ \partial \ell_k}  { \partial s_k}$ to the center\;
\end{algorithm}

Upon receipt, the central server executes Alg.~\ref{algorithm:center}: it accumulates the gradients across schools. A BFGS update then adjusts $\bm{\alpha}$ and $\bm{\beta}$, as well as each $s_k$. For identification, we impose the constraint $\sum_{k=1}^K s_k = 0$ by minus the average school effect from each $s_k$ after every update. Convergence is assessed by verifying that all absolute gradients fall below the tolerance $\epsilon$. If convergence is not achieved, the updated parameters are redistributed and the cycle repeats. This interplay of local E-steps and a central M-step continues until parameter estimates stabilize, ensuring a scalable, communication-efficient solution for federated IRT modeling. 

\begin{algorithm}[!htb]
    \caption{FedIRT---Estimation algorithm in center}
    \label{algorithm:center}
    \SetKwInOut{Given}{Given}
    \Given{Tolerance for convergence $\epsilon$}
    \KwIn{Gradients:  $\frac{\partial \ell_k}{\partial \alpha_j}, \frac{\partial \ell_k}{\partial \beta_j}, j=1,\dots,J$ and $ \frac{\partial \ell_k}{\partial s_k}, k=1,\dots,K$}
    \KwOut{Estimates for item parameters $\bm{\alpha}$ and $\bm{\beta}$ and school effect $\bm{s}$}
    
    Initialize parameters: set all $\alpha_j = 1$ and $\beta_j=0$ for $j=1,\dots, J$ and all $s_k = 0$ for $k=1,\dots,K$\;
    \While{not converged}{
        \For{school $k=1$ \KwTo $K$}{
            Send updated item parameters $\bm{\alpha}$ and $\bm{\beta}$ and school effect $s_k$ to the school\;
            Receive gradients $\frac{\partial \ell_k}{\partial \alpha_j}, \frac{\partial \ell_k}{\partial \beta_j}, j=1,\dots,J$ and $\frac{\partial \ell_k}{\partial s_k}$\;
            Sum up gradients across schools\;
            Update parameters by BFGS Method\;
            identification constraint: $s_k = s_k - \text{mean}(\bm{s})$\;
            Check convergence criteria: $\frac{\partial \ell_k}{\partial \alpha_j}<\epsilon, \frac{\partial \ell_k}{\partial \beta_j}<\epsilon, \frac{\partial \ell_k}{\partial s_k}<\epsilon,$\;
        }
    }
\end{algorithm}
% In practice, it's possible that extreme response patterns (e.g., most students fail to answer an item correctly) may lead to instability in the parameter updates. Some techniques can be introduced to mitigate this issue, such as truncated mean. In distributed machine learning and Federated Learning, robust aggregation is crucial for defending against outliers, which may stem from non-IID data, system faults, or malicious Byzantine attacks \citep{fang2020local}. The Truncated Mean (or Trimmed Mean) is a widely adopted robust aggregation strategy that ensures stability by discarding extreme values \citep{yin2018byzantine}. Given a set of $N$ gradient vectors $\{\bm{g}_1, \ldots, \bm{g}_N\}$ from clients, the Truncated Mean operates on each parameter dimension $j$. For the set of scalar values $\{g_{1j}, \ldots, g_{Nj}\}$, let $g_{(i)j}$ denote the $i$-th order statistic (i.e., the $i$-th smallest value). The aggregator first defines a trim parameter $r$ (where $2r < N$). The Truncated Mean $\hat{g}_j$ for dimension $j$ is then defined as the average of the ``central'' values, formally:
% \begin{equation}
%     \hat{g}_j = \frac{1}{N - 2r} \sum_{i=r+1}^{N-r} g_{(i),j}
% \end{equation}
% By systematically removing the $r$ smallest and $r$ largest values from the aggregation, this method prevents outliers from disproportionately influencing the global model update, thus ensuring robustness. In our implementation, the $r$ is defined by the percentage of truncated, e.g., truncated means ($10\%$) $r = \lfloor 0.1N \rfloor$.

Although the above description is based on the 2PL model, it can be extended to the Partial Credit Model by appropriately modifying the log-likelihood and gradient computations. The details are omitted here. We have developed an \texttt{R} package, \texttt{FedIRT}, to implement this algorithm for both the 2PL IRT model and the PCM.

\subsection{Differentially Private FedIRT (Student-Level DP with Central Gaussian Noise)}
In FedIRT, to prevent the leakage of individual students' response data from each school, one safeguard is imposed by transmitting only aggregated gradients rather than raw data. However, this measure alone may not fully protect against inference attacks that exploit these aggregated statistics. To enhance more robust privacy mechanisms, we integrate student-level Differential Privacy into the FedIRT and develop FedIRT-DP extension. This integration ensures that the inclusion or exclusion of any single student's data has a limited impact on the overall model updates, thereby providing stronger privacy guarantees.

We extend FedIRT with student-level DP using a clipped, per-student gradient mechanism with central Gaussian noise to replace two steps in the basic FedIRT, as shown in the dash-line blocks of Fig.~\ref{fig:flowchart}. Each school computes per-student contributions to the 2PL log-likelihood score, clips their $\ell_2$ norm to a global constant $C$, and sums them locally (Alg.~\ref{alg:dp_school}). The server then adds Gaussian noise to the aggregated vector before updating parameters via another numerical method Adam \citep{kingma2014adam}, which is designed for stochastic gradient descent with subsampling (Alg.~\ref{alg:dp_center}). In order to improve model identification and stability, we transform the discrimination parameters via $\alpha_j=\exp(a_j)$ and optimize over the unconstrained $a_j$. To further stablize noisy updates, independent MAP priors are imposed on all parameters, which acts as ridge regularization.

\paragraph{Per-student gradients.}
For school $k$ with $N_k$ students and $J$ items, once the center server broadcasts current parameters $a_j$, $\beta_j$, and $s_k$, each school computes per-student gradients using Eqs.~\ref{eq:grad_alpha}, \ref{eq:grad_beta}, and \ref{eq:grad_school_effect}. Stacking all gradients for student $i$ in school $k$ gives
\begin{equation}
    \bm{g}_{ik} = \left(g_{ik}^{(a_1)},\dots,g_{ik}^{(a_J)},g_{ik}^{(\beta_1)},\dots,g_{ik}^{(\beta_J)},g_{ik}^{(s_k)}\right)\in\mathbb{R}^{2J+1}.
\end{equation}

\paragraph{Subsampling and clipping.}
We perform independent Poisson subsampling of students at each school with rate $q_s\in(0,1)$\footnote{Each student is included in the subsample independently with probability $q_s$ (a Bernoulli($q_s$) draw for each student), so the total number selected is random.
}; for each sampled student, we clip the contribution to $\|g_i\|_2\le C$,
\begin{equation}
    \tilde{\bm{g}}_{ik} = \bm{g}_{ik}\cdot \min\left(1,\frac{C}{\|\bm{g}_{ik}\|_2}\right).
\end{equation}
and sum within the school: $\bm{G}_k=\sum_{i\in\mathcal{S}_k}\tilde g_i$, where $\mathcal{S}_k$ is the set of sampled students in school $k$.

\paragraph{Secure aggregation and central noise.}
All schools use a secure-aggregation protocol (standard in federated learning) so the server only sees the sum of local gradients $\bm{G}=\sum_{k=1}^K \bm{G}_k$. The central server then adds i.i.d.\ Gaussian noise $\bm{z}\sim\mathcal{N}(\bm{0},\sigma^2 C^2 I)$ to obtain $\tilde{\bm{G}} = \bm{G} + \bm{z}$.

We use student-level DP with the central Gaussian mechanism at level $(\sigma,\delta)$, where $\delta=10^{-6}$ in all experiments. (A formal $\varepsilon$ bound, which indicates the magnitude of the privacy loss, may be obtained by standard Rényi Differential Privacy \citep[RDP;][]{mironov2017renyi} accounting under Poisson subsampling. See more details in Appendix.)

\paragraph{MAP-Adam updates and projections.}
For $(a_1,\dots,a_J,\beta_1,\dots,\beta_J,s_1,\dots,s_K)$,  independent Gaussian priors
$a_j\sim\mathcal{N}(0,\tau_\alpha^2)$, $\beta_j\sim\mathcal{N}(0,\tau_\beta^2)$, $s_k\sim\mathcal{N}(0,\tau_s^2)$. We then add MAP penalities to the log-likelihood gradient:
\begin{equation}
    \bm{g}_{\mathrm{MAP}} = \tilde{\bm{G}} - \left(\frac{\bm{a}}{\tau_\alpha^2},\frac{\bm{\beta}}{\tau_\beta^2},\frac{\bm{s}}{\tau_s^2}\right).
\end{equation}
We then take an Adam step with per-parameter learning rates $(\eta_\alpha,\eta_\beta,\eta_s)$.
During the first 10 rounds, we softly clamp $\alpha_j=\exp(a_j)$ to $[0.2,3.0]$ to stabilize training, and we re-center $\bm{s}$ each round to enforce $\sum_k s_k=0$. We maintain a Polyak running average over the last $W=10$ rounds for reporting, mitigating the bias and extra variance introduced by DP noise. More hyperparameter details are in Appendix.

\begin{algorithm}[!htb]
\caption{FedIRT-DP---DP gradient computation algorithm in school $k$, $k=1,\dots,K$}
\label{alg:dp_school}
\KwIn{$X_k$; bins $V(n),A(n)$ for $n=1,\dots,q$; current $\alpha_j=\exp(a_j),\beta_j,s_k$; DP hyper-params $q_s,C$}
\KwOut{Masked share of $\bm{G}_k$ for secure aggregation}
\BlankLine
Compute $\pi_{ijk}(n)$ for $i=1,\dots,N_k$, $j=1,\dots,J$\;
$G_k\leftarrow \mathbf{0}$ in $\mathbb{R}^{2J+1}$\;
\For{$i=1$ \KwTo $N_k$}{
    \tcp{Poisson-subsample students with probability $q_s$ (privacy amplification)}
    \If{Bernoulli$(q_s)=0$}{\textbf{continue}}
    Compute bin posterior probabilities $p_{ik}(n)$ for $n=1,\dots,q$\ according to Eq.~\ref{eq:bin_posterior}\;
    \For{$j=1$ \KwTo $J$}{
        $\displaystyle g_{ik}^{(a_j)} \leftarrow \alpha_j\sum_{n=1}^q \!\big(V(n)+s_k-\beta_j\big)\big(x_{ij}-\pi_{ijk}(n)\big)p_{ik}(n)$\;
        $\displaystyle g_{ik}^{(\beta_j)} \leftarrow -\alpha_j \sum_{n=1}^q \!\big(x_{ij}-\pi_{ijk}(n)\big)p_{ik}(n)$\;
    }
    $\displaystyle g_{ik}^{(s_k)} \leftarrow \sum_{j=1}^J \alpha_j \sum_{n=1}^q \!\big(x_{ij}-\pi_{ijk}(n)\big)p_{ik}(n)$\;
    $\bm{g}_{ik} \leftarrow \big(g_{ik}^{(a_1)},\dots,g_{ik}^{(a_J)},g_{ik}^{(\beta_1)},\dots,g_{ik}^{(\beta_J)},g_{ik}^{(s_k)}\big)$\;
    \tcp{User-level DP clipping, $\lVert \cdot \rVert_2$ is the $\ell^2$ (Euclidean) norm of the vector}
    $\bm{g}_{ik} \leftarrow \bm{g}_{ik} \cdot \min\!\left(1,\frac{C}{\lVert \bm{g}_{ik}\rVert_2}\right)$\;
    $\bm{G}_k \leftarrow \bm{G}_k + \bm{g}_{ik}$\;
}
\textbf{SecureAgg.sendMaskedShare}$(\bm{G}_k)$ \tcp*{only masked shares are sent}
\end{algorithm}

\begin{algorithm}[!htb]
\caption{FedIRT-DP---Estimation algorithm in center}
\label{alg:dp_center}
\KwIn{Tolerance $\epsilon$; DP hyper-params $(R,q_s,C,\sigma,\delta)$; priors $(\tau_a,\tau_\beta,\tau_s)$; learning rate $(\eta_a,\eta_\beta,\eta_s)$}
\KwOut{Estimates $(\bm{\alpha},\bm{\beta},\bm{s})$; privacy guarantee $(\varepsilon,\delta)$}
\BlankLine
Initialize $\alpha_j\!=\!1$ (so $a_j\!=\!0$), $\beta_j\!=\!0$, $s_k\!=\!0$, $j=1,\dots,J$, $k=1,\dots,K$\;
\For{$t=1$ \KwTo $R$}{
    Broadcast $(\bm{\alpha},\bm{\beta},\bm{s})$ to all schools\;
    $\bm{G} \leftarrow$ \textbf{SecureAgg.receiveSum()} \tcp*{gets $\sum_k \bm{G}_k$ only}
    $\bm{z} \sim \mathcal N(\mathbf{0},\sigma^2 C^2 I)$;\quad $\widehat{\bm{G}} \leftarrow \bm{G} + \bm{z}$ \tcp*{central Gaussian mechanism}
    \tcp{Add MAP priors (ridge)}
    $\bm{g}_{\mathrm{MAP}} \leftarrow \widehat{\bm{G}} - (\bm{\alpha}/\tau_a^2,\;\bm{\beta}/\tau_\beta^2,\;\bm{s}/\tau_s^2)$\;
    \tcp{Update parameters by Adam}
    $(\bm{\alpha},\bm{\beta},\bm{s}) \leftarrow \text{Adam} (\bm{g}_{\mathrm{MAP}}, m_{t-1}, v_{t-1},\beta^{(\text{Adam})}_1=0.9,\beta^{(\text{Adam})}_2=0.999, \eta_a,\eta_\beta,\eta_s)$\;
    $\alpha_j \leftarrow \exp(a_j)$; \; if $t\le10$: $\alpha_j\leftarrow \mathrm{clip}(\alpha_j,0.2,3.0)$\;
    $\bm{s} \leftarrow \bm{s} - \mathrm{mean}(\bm{s})$ \tcp*{identifiability}
    Maintain Polyak average over last 10 rounds\;
    \If{$\|\bm{g}_{\mathrm{MAP}}\|_\infty < \epsilon$ \textbf{and} relative param change $< \epsilon$}{\textbf{break}}
}
\tcp{Privacy accounting (see Appendix for details)}
Let $R^*$ be the number of rounds actually executed\;
Compose per-round RDP over $R^*$ rounds: $\hat\varepsilon_{\mathrm{total}}(\alpha)=\sum_{t=1}^{R^*}\varepsilon_t(\alpha;\,q_s,\sigma)$\;
$\varepsilon^*\leftarrow\min_{\alpha>1}\bigl\{\hat\varepsilon_{\mathrm{total}}(\alpha)-\ln\delta/(\alpha-1)\bigr\}$\;
\Return Polyak-averaged $(\alpha,\beta,s)$; privacy guarantee $(\varepsilon^*,\delta)$
\end{algorithm}

\section{Simulation Study for FedIRT and FedIRT-DP}
To assess the efficacy of the basic FedIRT framework and FedIRT-DP extension, we conducted three simulation studies. The first study compares FedIRT and FedIRT-DP with established IRT packages under the standard 2PL model. The second evaluates the impact of treating school effects as fixed versus random. The third investigates the robustness of our methods when extreme response patterns are present.

\subsection{Study 1: Equivalence to \texttt{ltm} and \texttt{mirt}}

The first simulation aims to compare our algorithms with established methods. In order to directly compare the performance of our methods against two commonly used R packages for IRT estimation, \texttt{ltm} \citep{rizopoulos2007ltm,baker2004item} and \texttt{mirt} \citep{chalmers2012mirt, partchev2009irtoys}, which do not account for school effects, we set and fixed all school effects to zero ($s_k = 0$) in the data generating process and for our FedIRT and FedIRT-DP implementations.

\subsubsection{Data Generating Process}
We developed a data generation procedure involving specified numbers of items ($J$), schools ($K$), and students within each school ($N_k$). For each simulation scenario, we fixed the number of items ($J=10$), the number of schools ($K=10$), and varied student sample sizes ($N_k \in \{50, 100, 300\}$). In each scenario, item parameters ($\alpha_j$ and $\beta_j$) and student abilities ($\theta_i$) were first generated systematically and fixed. The response matrix was generated repeatedly using fixed item parameters ($\bm{\alpha}$ and $\bm{\beta}$) to rigorously assess the algorithm's consistency. 

We employed distinct parameter ranges to ensure robustness in our model. The high-discrimination condition set $\alpha \in [1,2]$, while the low-discrimination condition set $\alpha \in [0.5,1]$. Similarly, the high-difficulty condition employed $\beta \in [0,1]$, and the low-difficulty condition utilized $\beta \in [-1,0]$. Parameters $\alpha_j$ and $\beta_j$ followed uniform distributions within these specified ranges, whereas student abilities $\theta_i$ were sampled from a standard normal distribution, $\theta_i \sim \mathcal{N}(0,1)$.

Responses were generated according to the 2PL model probability defined in Eq.~\ref{eq:2PL}, setting $s_k=0$. Specifically, responses were drawn from a Bernoulli distribution based on the computed probabilities. We repeated the simulation process 100 times ($T=100$) to examine the robustness of our method.

\subsubsection{Estimating Global Parameters using Different Algorithms}
We compared our framework FedIRT and FedIRT-DP extension with MMLE algorithm implemented in R package \texttt{ltm}, and \texttt{mirt} to estimate the global model parameters, $\bm{\alpha}$ and $\bm{\beta}$. For the two popular IRT packages, we considered two approaches to estimating the item parameters. The first approach was that we estimated item parameters as if the data was aggregated from all schools. This is \textit{centralized estimation} because all computation are conducted in the center. The other approach utilized a \textit{meta-analysis} function, similar to the Eq.~\ref{eq:fedavg_param} for FedAvg. The meta-analysis algorithm also has the advantage of protecting students' privacy, because it does not need to upload the original response patterns to the center. In each school $k$, the \texttt{mirt} package was used to estimate the discriminations and difficulties with local dataset and send the estimates $\alpha_{jk}$ and $\beta_{jk}$ to the center. Then, the center takes a weighted average to represent the global parameters \citep{borenstein2010basic}, which is shown in the following formula: for $j=1,\dots,J$,
\begin{equation}\label{eq:meta}
    \hat{\alpha}_j = \sum\limits_{k=1}^{K}\frac{ N_k  }{N} \alpha_{jk}, \hat{\beta}_j = \sum\limits_{k=1}^{K}\frac{ N_k  }{N} \beta_{jk}
\end{equation}

\subsubsection{Evaluation Criteria}
We adopted Mean Squared Error (MSE) and Bias as metrics for performance evaluation. Mean Squared Error \citep{geman1992neural} is a widely used way to measure differences between estimation results and real values of data, We use superscript $(t)$ to denote estimation results based on the $t$-th response set for $t=1,\dots,T$. The MSE is defined as follows:
\begin{equation}
    \text{MSE}_{\alpha} = \frac{ \sum\limits_{t=1}^{T} \sum\limits_{j=1}^J ( \hat{\alpha}_{j}^{(t)} - \alpha_{j} )^2 }{JT},\quad
    \text{MSE}_{\beta} = \frac{ \sum\limits_{t=1}^{T} \sum\limits_{j=1}^J ( \hat{\beta}_{j}^{(t)} - \beta_{j} )^2 }{JT}
\end{equation}

Bias is another widely used way to measure how the estimation results deviate from the true values \citep{friedman2000additive}. In our study, bias is defined by removing the square in the MSE formula.

\subsubsection{Results of Study 1}
\begin{figure}[!htb]
    \centering
    \centerline{\includegraphics[width=\linewidth]{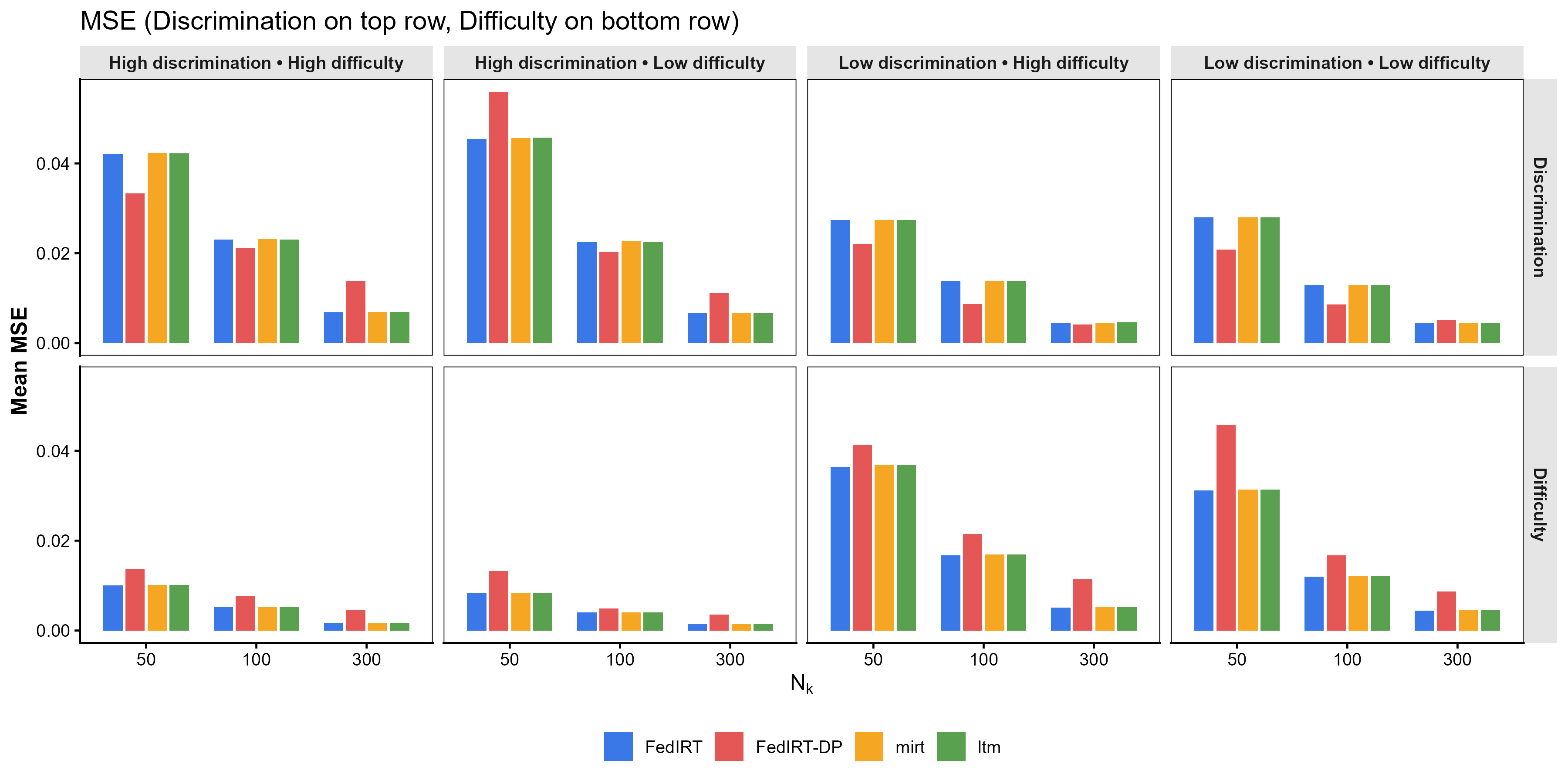}}
    \caption{Comparison of MSE for FedIRT, FedIRT-DP centralized estimation by \texttt{ltm} and \texttt{mirt} across different sample sizes and levels of item discrimination and difficulty.}
    \label{fig:study1_mse_plot}
\end{figure}

We first compared our proposed methods with the centralized estimation results obtained using \texttt{ltm} and \texttt{mirt}. As shown in Fig.~\ref{fig:study1_mse_plot}, across all conditions the mean MSE drops steadily as the school size increases from $N_k=50$ to $N_k=300$.  FedIRT (blue) tracks the centralized baselines (\texttt{mirt}, \texttt{ltm}; orange/green) almost exactly for both discrimination and difficulty.  FedIRT-DP (red) shows the largest errors when $N_k=50$, with the gap most visible for discrimination in the high-discrimination and low difficulty setting; the gap narrows at $N_k=100$ and is small by $N_k=300$, where all methods yield MSEs close to zero.  MSEs are generally higher in the high-discrimination panels (top row, first two columns) and lower in the low-discrimination panels, consistent with steeper item slopes being harder to estimate under noise.  The general higher MSEs for FedIRT-DP follows directly from the mechanism: per-user gradient clipping reduces large item-slope updates, the added Gaussian noise lowers the signal-to-noise ratio, and the MAP penalty further damps updates.  With more students per school (or, equivalently, more effective information), the data term dominates and the DP method approaches the non-DP fits.

\begin{figure}[!htb]
\centering
\centerline{\includegraphics[width=\linewidth]{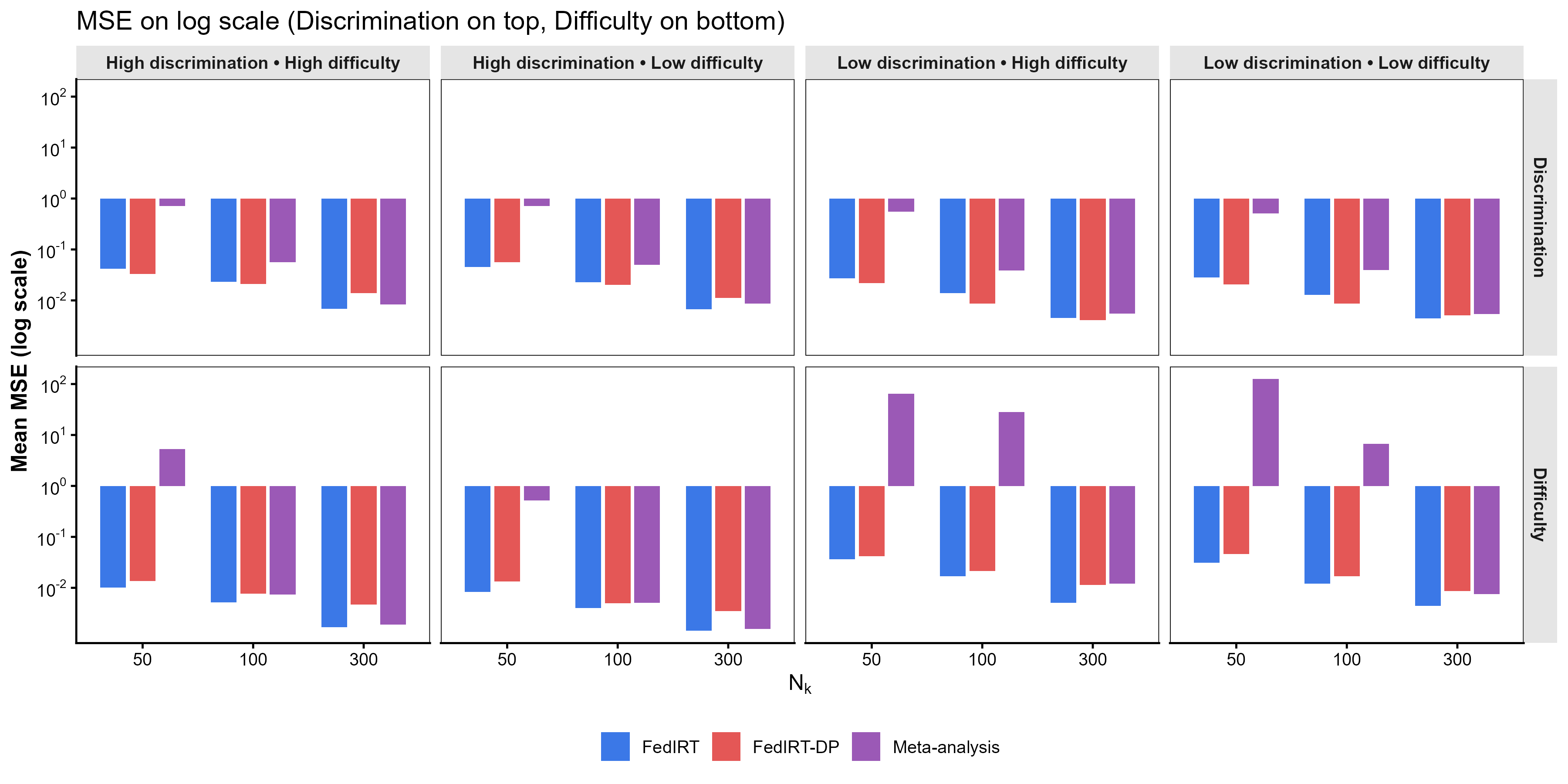}}
\caption{Comparison of MSE for FedIRT, FedIRT-DP and meta-analysis approach by \texttt{mirt} across different sample sizes and different levels of item discrimination and difficulty.}
\label{fig:study1_mse_meta_plot}
\end{figure}

We subsequently compared our methods with the meta-analysis approach described in Eq.~\ref{eq:meta}, which uses local estimates obtained via the \texttt{mirt} package. Due to the significantly higher MSE of the meta-analysis results, we adopted a logarithmic scale for the y-axis in Fig.~\ref{fig:study1_mse_meta_plot} to facilitate comparison. Mean MSE declines with larger school size in every panel, but the meta-analysis (purple) remains clearly worse than the federated estimators (FedIRT and FedIRT-DP, blue/red). At $N_k=50$ its errors are often one-two orders of magnitude larger for both discrimination and difficulty; the gap narrows at $N_k=100$ and $N_k=300$ but persists across most truth settings. This pattern arises because the meta approach fits each school in isolation and then averages item parameters. That averaging discards cross-school pooling and fails to align difficulties to a common anchor. Small-sample variability from each school therefore accumulates rather than cancels. In contrast, the federated estimators aggregate gradients for a single joint likelihood, so information is pooled across schools and parameters are estimated on a common scale, which yields substantially smaller MSE.

Mean bias for item discrimination $\alpha$ and difficulty $\beta$ is negligible for FedIRT, \texttt{ltm}, and \texttt{mirt} across all design conditions: $\alpha$-bias lies roughly between $-0.01$ and $0.03$, and $\beta$-bias is typically within $\pm 0.02$. By contrast, the meta-analysis approach exhibits much larger positive bias for $\alpha$, especially when the per-site sample is small: at $N_k=50$ the mean $\alpha$-bias is about $0.34$--$0.43$, decreasing to $0.07$--$0.12$ at $N_k=100$ and to roughly $0.02$ at $N_k=300$. Its $\beta$-bias is smaller but can reach about $0.16$ in the most extreme case. FedIRT-DP shows average bias for both $\alpha$ and $\beta$ comparable in magnitude to FedIRT once $N_k\ge 100$, with some noticeable departures from zero at $N_k=50$. These signs are consistent with the update mechanics: clipping and noise attenuate the data gradients that grow with the discrimination, while the Gaussian prior on $\alpha=\log a$ pulls estimates toward its mean (so $a$ toward 1).  The optimizer then compensates by shifting $\beta$ to preserve observed success rates\footnote{Mechanistically, in the 2PL log-likelihood the gradient with respect to $\alpha_j$ is typically negative when $a_j$ is already large (it pushes the slope down), but clipping and noise attenuate this negative signal toward zero, so the optimizer takes smaller downward steps and stops at a larger $\alpha_j$ than the non-DP fit; because $P_{ijk}=\sigma\!\big(a_j(\theta_{ik}+s_k-\beta_j)\big)$, the increase in $a_j$ is compensated by moving $\beta_j$ closer to the center of the ability distribution so that $a_j(\theta_{ik}+s_k-\beta_j)$, and hence the predicted probabilities, remain approximately unchanged.}.

% We also compared the bias of the estimates obtained using our methods with those from the meta-analysis approach. The results are presented in Fig.~\ref{fig:study1_bias_meta_plot}. Bias for the federated estimators is close to zero across sample sizes and truth settings, while the meta-analysis shows sizable positive bias-most pronounced for discrimination at $N_k=50$ and still visible at $N_k=300$; difficulty exhibits the same direction in several cells. The explanation follows the same logic as for MSE: estimating each school separately and averaging parameters without proper linking introduces scale drift and non-invariant averaging, which produces systematic offsets instead of cancelling estimation error.
% \begin{figure}[!htb]
% \centering
% \centerline{\includegraphics[width=\linewidth]{fig/study1_bias_methods_meta.png}}
% \caption{Comparison of bias for FedIRT, FedIRT-DP and meta-analysis approach by \texttt{mirt} across different sample sizes and different levels of item discrimination and difficulty.}
% \label{fig:study1_bias_meta_plot}
% \end{figure}

\subsection{Study 2: School Effect as Fixed Effect}
In Study 1, in order to directly compare with the established methods, we set the school effect $s_k = 0$ for each school $k$, thus treating school effects implicitly as random. Under these conditions, the estimation of a school's ability typically involves calculating the average of all individual student ability estimates within that school. In the current study, we demonstrate through simulation that explicitly estimating school effects as fixed effects, jointly with item parameters, enhances the accuracy of school effect estimates compared to the conventional approach of treating these effects as random, while maintaining reliable item parameter estimates.

We considered a scenario with five schools ($K = 5$) and defined a fixed vector of school effects as $\bm{s} = (-1, -0.5, 0, 0.5, 1)$. Each school had an equal number of students ($N_k = 100$). We generated 10 items ($J = 10$) with item discrimination parameters ($\alpha_j$) and item difficulty parameters ($\beta_j$) randomly sampled from uniform distributions, specifically $\alpha \in [0.5, 2]$ and $\beta \in [-1, 1]$. The simulation procedure was identical to Study 1 but with school effects were explicitly assigned as $\bm{s}$. We conducted the simulation over 100 replications ($T = 100$). Two approaches were compared: (1) treating school effects as fixed effects and estimating them jointly with item parameters using our FedIRT and FedIRT-DP framework; (2) treating school effects as random effects, where $s_k$ in the model is set to $0$ for all schools, and school effects were estimated by averaging individual student abilities within each school.

\begin{figure}[!htb]
\centering
\centerline{\includegraphics[width=\linewidth]{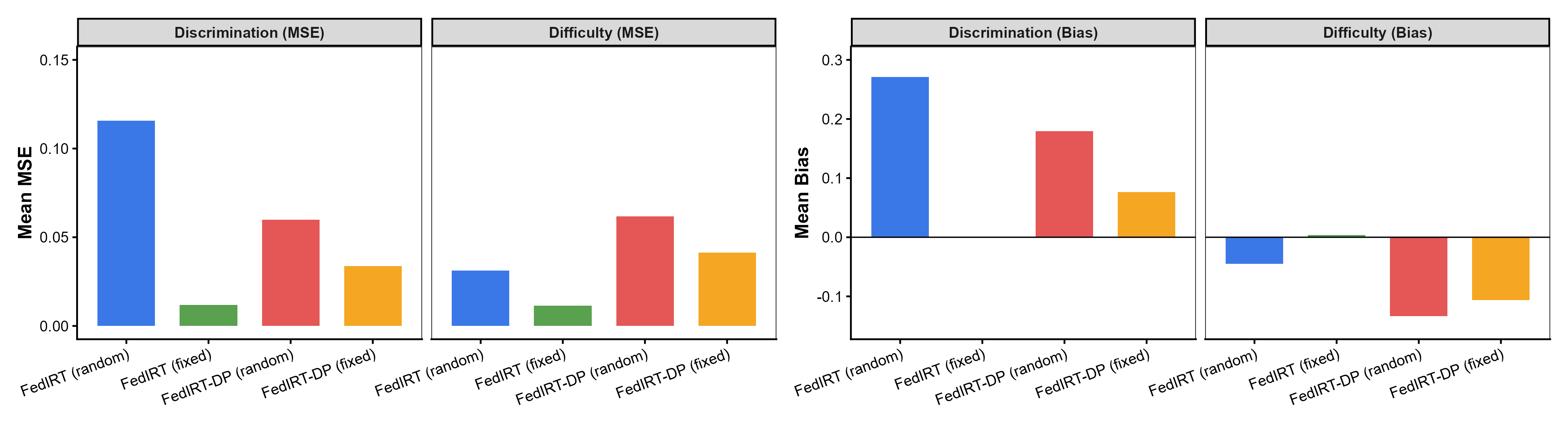}}
\caption{Comparison of MSE and bias for item parameter estimation between fixed and random school effect approaches.}
\label{fig:study2_item_mse_bias}
\end{figure}

First, we evaluated the MSE and bias for item parameter estimates derived from both approaches (fixed versus random school effects). Fig.~\ref{fig:study2_item_mse_bias} clearly illustrates that when school effects are treated as fixed and estimated jointly with the items, both FedIRT and FedIRT-DP deliver smaller errors for the item parameters than the ``random-effect'' alternative that derives school effects by averaging student abilities. In the discrimination panel, the fixed-effect variants reduce MSE substantially and bring the mean bias close to zero, whereas the random-effect variants show noticeably larger MSE and a clear positive bias. A similar pattern holds for difficulty: the fixed-effect fits lower MSE and keep bias near zero, while the random-effect fits, especially with DP, exhibit larger MSE and a negative bias. This occurs because leaving $s_k$ implicit forces school-level shifts to be absorbed partly by the item parameters; the estimated discrimination and difficulty then compensate for between-school location shifts, inflating error. Making $s_k$ an explicit parameter allocates that systematic shift to the school term, stabilizing the item updates and removing leakage from $s_k$ into $\alpha_j$ and $\beta_j$. Any remaining gap under DP reflects clipping and injected Gaussian noise, which raise variance and introduce mild shrinkage, but the fixed-effect formulation still recovers cleaner item estimates than the random-effect alternative.

\begin{figure}[!htb]
\centering
\centerline{\includegraphics[width=\linewidth]{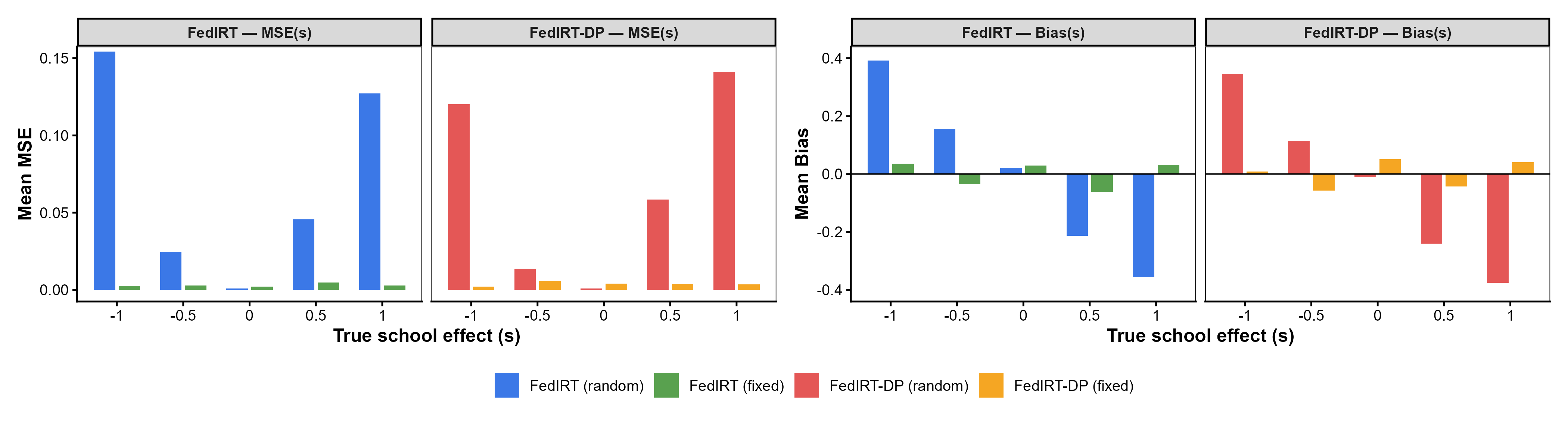}}
\caption{Comparison of MSE and bias for school effect estimation between fixed and random school effect approaches.}
\label{fig:study2_schools_mse_bias}
\end{figure}

Subsequently, we assessed the accuracy of school ability estimation by comparing MSE and bias across approaches. As shown in Fig.~\ref{fig:study2_schools_mse_bias}, The random-effect approach yields a pronounced U-shape in MSE across the true values $\bm{s}$: errors are largest at the extremes and smallest near zero. Its bias is correspondingly signed. Schools with negative truth are overestimated and those with positive truth are underestimated, showing clear shrinkage toward zero. By contrast, modelling $s_k$ as fixed produces near-flat, low MSE across all five values and biases close to zero for both FedIRT and FedIRT-DP. The mechanism is straightforward: averaging student posteriors to form $\hat{s}_k$ mixes school location with item misspecification and inherits shrinkage from the ability prior and identifiability constraints, so extreme schools are pulled toward the grand mean. Joint estimation treats $s_k$ as a parameter in the likelihood, so the model assigns the location shift to the school rather than to the items or abilities, which removes the shrinkage and reduces variance.

\subsection{Study 3: Extreme Response Patterns}
While our FedIRT estimator matches standard software under ideal conditions, it can be sensitive to \emph{extreme response rows} that occur in practice (e.g., students who answer every item incorrectly or correctly). The FedIRT-DP extension moderates such rows via user-level differentially private aggregation: per-user $\ell_2$ clipping and central Gaussian noise reduce the leverage of any single user. To assess robustness, we constructed two contamination scenarios for each school's response matrix $\bm{x}_{ik}$: (1) \textit{zeros}: a subset of students respond $0$ to all items; (2) \textit{ones}: a subset respond $1$ to all items. We computed MSE and bias as in Study~1. Each study used $K=10$ schools with $N_k=100$ students and $J=10$ items, and we ran $T=100$ replications. The proportion of extreme rows in each school varied from $0.10$ to $0.50$.

\begin{figure}[!htb]
\centering
\centerline{\includegraphics[width=\linewidth]{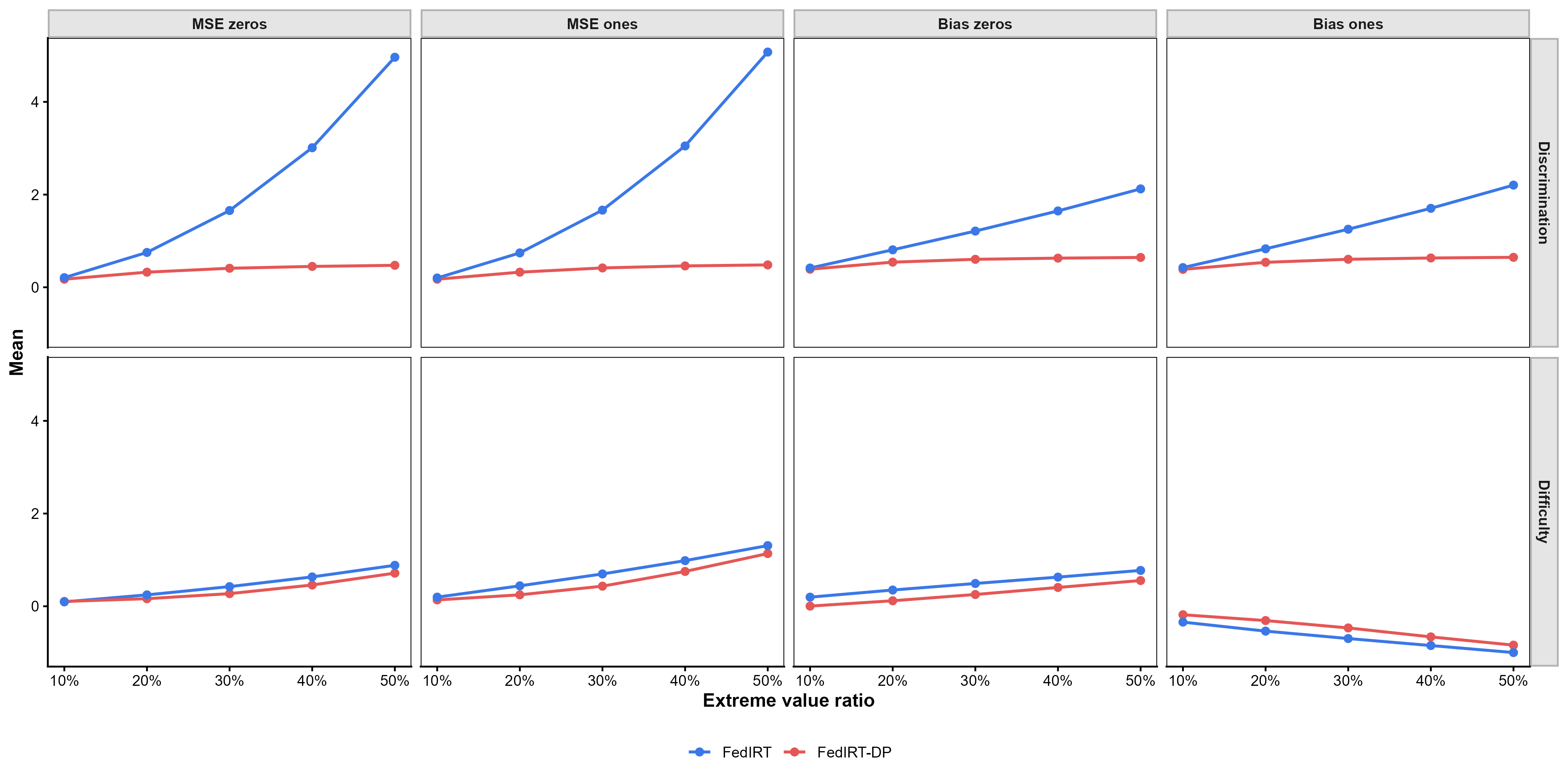}}
\caption{Comparison of MSE and bias between basic FedIRT framework and FedIRT-DP extension across different rates of extreme responses.}
\label{fig:study3}
\end{figure}

For discrimination (top row of Figure~\ref{fig:study3}), both MSE and bias increase monotonically with the proportion of extreme rows. Under FedIRT, the increase is steep, especially in the \textit{ones} scenario, and the bias grows steadily. Under FedIRT-DP, the curves are markedly flatter: MSE rises slowly and remains well below the non-DP values across all contamination levels, and the bias increases only slightly. For difficulty (bottom row), contamination has a milder effect: MSE grows gradually in both scenarios, with FedIRT-DP consistently a bit lower than FedIRT; the bias drifts downward as contamination increases, but the DP estimator stays closer to zero. These patterns follow from the update mechanics. Extreme rows generate outlying per-user gradients that can dominate the school aggregation; FedIRT aggregates these unboundedly, inflating variance and shifting parameters in the direction implied by the extremes (all-$0$ or all-$1$). FedIRT-DP clips each user's gradient before aggregation and adds Gaussian noise at the center, so the influence of extreme rows is capped and partially randomized. As contamination increases, the DP curves do rise (information is lost when many rows are uninformative), but the growth is modest compared with the non-DP estimator, yielding more robust item-parameter estimates.

\section{Empirical Study with Real-World School Data}
We also provide an empirical illustration of our proposed FedIRT framework and FedIRT-DP extension using real data that are naturally partitioned by institutions (schools). We analyzed a small subset of the TIMSS 2019 International Database \citep{fishbein2021timss} consisting of 50 schools and 15 dichotomous items. Per-school sample sizes were small (approximately 6-8 students), and the dataset contained no missing responses.

\begin{figure}[!htb]
\centering
\includegraphics[width=\linewidth]{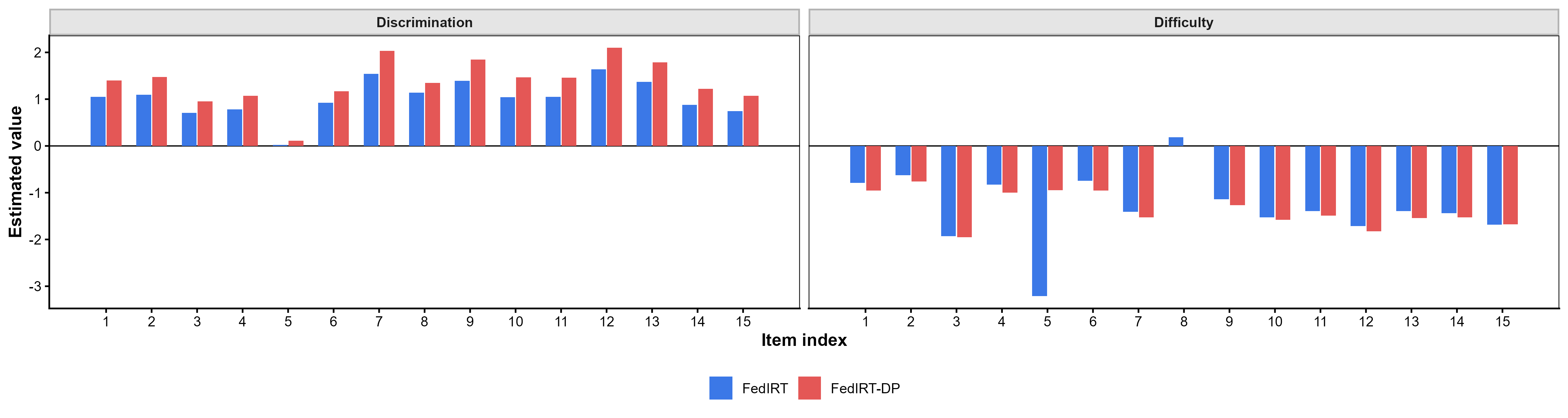}
\caption{Estimated item parameters from FedIRT and FedIRT-DP.}
\label{fig:emp_items}
\end{figure}

Figure~\ref{fig:emp_items} shows the 15 item estimates. For \emph{discrimination} (left panel), estimates span roughly \(0.7\)--\(1.6\) under FedIRT and \(0.9\)--\(2.1\) under FedIRT-DP. For \emph{difficulty} (right panel), most items are hard with values between about \(-0.6\) and \(-3.2\). The two methods preserve the same ranking of most items. FedIRT-DP tends to return higher discriminations and less negative difficulties, a pattern consistent with the DP update: per-user $\ell_2$ clipping and central Gaussian noise reduce the effective counter-gradient for large discrimination, nudging $\alpha_j$ upward; to match observed response rates the fitted thresholds $\beta_j$ shift toward zero.

\begin{figure}[!htb]
\centering
\includegraphics[width=\linewidth]{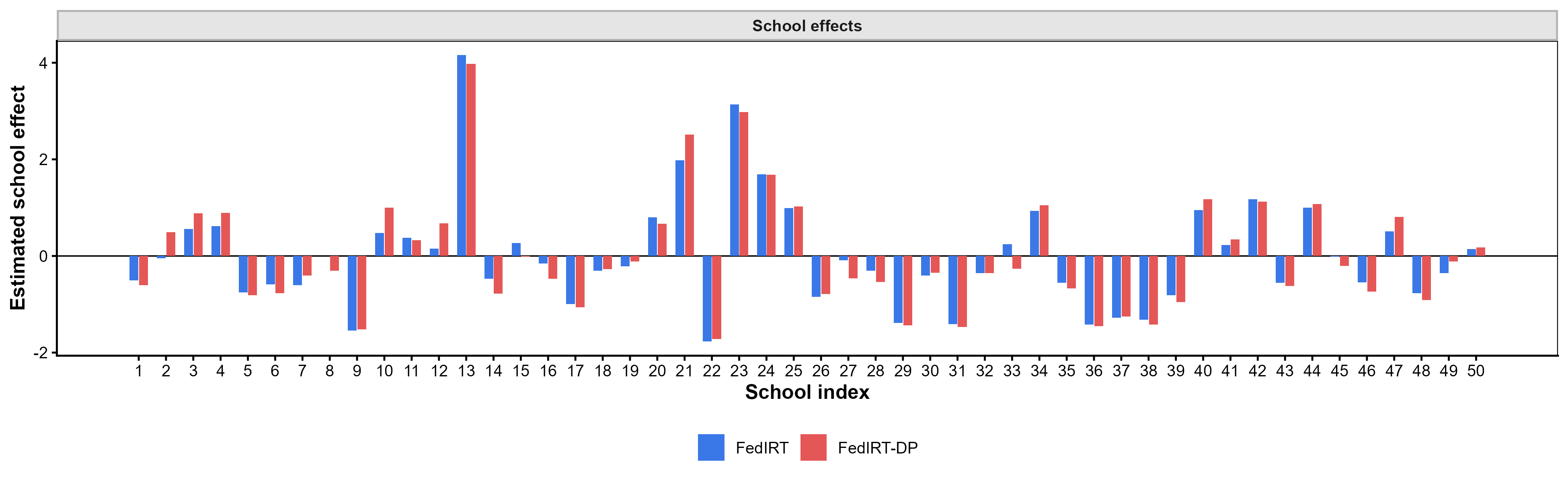}
\caption{Estimated school effects from FedIRT and FedIRT-DP. Positive values indicate higher expected performance on the latent scale after accounting for item parameters.}
\label{fig:emp_schools}
\end{figure}

Figure~\ref{fig:emp_schools} reports the 50 school effects on a common index scale (1--50). Estimates range from about \(-1.8\) to \(+4.2\) for FedIRT and \(-1.7\) to \(+4.0\) for FedIRT-DP. Across all schools the two profiles are very similar; differences are concentrated at the extremes, where FedIRT-DP is slightly closer to zero. This mild shrinkage is in line with the DP mechanism and the MAP regularization used in the DP aggregation.

Taken together, this empirical illustration demonstrates that the proposed framework and DP extension operate effectively on real data that are naturally partitioned by schools. It recovers item parameters in plausible ranges and produces school-level effects that are straightforward to interpret for substantive and policy discussions, all while preserving data locality.

\section{Potential Estimation Issues}
Some potential estimation issues common in item response models might affect the performance of our proposed FedIRT framework. Here we discuss some of these issues and possible solutions.
\subsection{Local Item Dependence (LID)}
Local item dependence (LID) arises when items share stimulus/context (e.g., testlets), violating conditional independence and inflating discrimination, reliability, and fit indices. In distributed settings, such dependence can differ by site (e.g., school-specific curricula), so naive federated aggregation can propagate biased gradients \citep{yen1984effects,christensen2017critical}. In practice, we recommend school-local LID screening (residual-correlation/$Q_3$ diagnostics) with site flags, followed by either testlet (bundle) modeling locally or down-weighting flagged item pairs before gradient formation; the center can then aggregate per-site “LID masks” to run sensitivity analyses.

\subsection{Nonnormal Latent Distributions}
Nonnormal latent distributions (skew/heavy tails) are common in applied settings and can degrade MML estimates and fit if one assumes standard normal ability, patterns that can differ across schools due to selection or tracking. Two pragmatic upgrades fit our federated loop: replace the fixed normal prior with a flexible mixture or empirical (Gaussian-Hermite) estimate at the center, or add a latent-distribution fit check and, if rejected, switch to mixture-prior updates while keeping item/school gradients unchanged.

Since the FedIRT algorithm is based on gradient descent, which is also the fundamental idea of deep learning for neural networks, we can also consider integrating deep learning techniques to enhance the flexibility of latent distribution estimation. IRT models can be transformed into neural networks and techniques such as variational autoencoders (VAEs) can be employed to capture complex latent structures \citep{urban2021deep,luo2025generative}. By incorporating these methods into the FedIRT framework, we can potentially improve the model's ability to handle nonnormal latent distributions while maintaining data privacy across institutions.

\subsection{Differential Item Functioning (DIF)}
Differential item functioning (DIF) is expected across sites when instruction, language, or context differ, and federated training can mask it if pooling is unchecked. Practically, maintain site (or site-cluster) contrasts by computing per-site item-gradient summaries and running center-side DIF screens (Mantel-Haenszel or IRT-LR based) before final global updates; items with consistent site-specific bias can be flagged for exclusion or multi-group calibration with anchored parameters \citep{holland2012differential}.

\subsection{Imbalanced Sample Size across Institutions}
When one site has a very large $N_k$ while others are small, its gradients dominate the federated update, so global item and shift estimates tend to align with that site's response distribution; this is a well-known issue in federated optimization under client heterogeneity/non-IID data and unbalanced participation \citep{li2019convergence}, and it is common in practice because institutional sizes vary widely. In IRT specifically, unequal sample sizes change the precision and small-sample bias of item parameters and school effects. Small-$N_k$ sites contribute high-variance gradients, so their local structure can be washed out in pooled updates; such imbalance across sites is therefore expected to vary and to affect calibration stability. To mitigate, weight site contributions toward effective information rather than raw counts (e.g., cap the per-round weight of the largest site; use group-balanced or stratified client sampling so each round draws a fixed number of small and large sites), and apply per-student clipping with the same $C$ so no site can dominate via a few extreme gradients. Also, report sensitivity analyses that reweight or temporarily exclude the largest site and compare item/school estimates; if substantive conclusions change, retain group-balanced sampling and shrinkage in the main analysis.

\section{Discussion}
Although the notion of federated learning has been discussed in many other fields, it has not been studied in the field of psychometrics. In the current study, we propose a novel methodological framework, Federated IRT, which integrates federated learning with traditional Item Response Theory models. FedIRT enables multiple institutions to collaboratively estimate item parameters and school-specific effects while preserving the confidentiality of individual student responses.  We have implemented this framework in an R package, \texttt{FedIRT}, supporting both the two-parameter logistic model and the partial credit model with a user friendly Shiny interface. We have also extended the basic FedIRT algorithm to incorporate user-level differential privacy and developed the novel FedIRT-DP method, providing formal privacy guarantees against potential information leakage during the federated estimation process.

Three simulation studies provide evidence for the proposed methods and the corresponding estimation procedures' effectiveness. In our first simulation study, the results from FedIRT matched the performance of popular established R packages (\texttt{ltm} and \texttt{mirt}) under a standard 2PL model based on pooled data, and it also surpassed the simple weighted-average local estimates produced by \texttt{mirt} across multiple schools. FedIRT-DP showed slightly higher MSE and bias, but its performance improved with larger per-school sample sizes, approaching that of FedIRT. The second simulation showed that modeling school effects as fixed parameters within FedIRT and FedIRT-DP yields more stable and precise estimates than treating these effects as random. Although FedIRT-DP performed slightly worse than FedIRT due to the introduction of noise and gradient clipping, in the third simulation study, it demonstrated superior robustness to extreme response patterns compared to the standard FedIRT algorithm. Finally, an empirical application provided additional evidence that FedIRT and FedIRT-DP delivers meaningful parameter estimates, all while preserving data privacy.

By combining IRT with federated learning and differential privacy, school boards and researchers can derive insights from geographically dispersed educational datasets without transferring raw response data. This approach not only strengthens data privacy but also motivates the development of more distributed and computationally efficient IRT modelling techniques. Such a framework is particularly advantageous in contexts governed by strict privacy regulations, including the current K-12, government, and higher education systems. Also, its use of distributed computation allows most processing occurring locally at each institution and reduces the load on any central server and fully leveraging available computational resources across sites \citep{han2019federated}. Moreover, it minimizes network bandwidth by transmitting only summary statistics per school in each iteration, rather than entire datasets.

FedIRT also facilitates standardized comparisons of student proficiency across diverse institutions without exposing individual responses. Traditionally, aligning scales across schools requires complex equating procedures to adjust for curricular and demographic differences. With FedIRT, a single metric representing each school's effect can be directly compared without data being transferred, yielding a more equitable assessment of student ability despite variations in instructional methods and student populations.

Furthermore, the differential privacy extension delivers a \emph{provable} limit on what can be learned about any individual student from the output, regardless of side information. The user-level formulation protects an entire response vector per student, which aligns with the unit of risk in educational testing. Two properties are especially important in practice: (i) post-processing immunity, which ensures that the guarantee is preserved by all downstream analyses of the released estimates, and (ii) composability, which allows institutions to track a cumulative privacy budget across training rounds and releases. Our FedIRT-DP instantiation operationalizes these benefits in IRT by (a) clipping per-student gradients to a fixed norm (bounding any one student's influence), (b) adding calibrated central Gaussian noise to the aggregate (ensuring an \((\varepsilon,\delta)\) guarantee), and (c) performing MAP updates that retain standard IRT interpretability. In addition to formal protection, the same clipping-and-noise mechanism improves robustness to atypical rows (e.g., all-zeros or all-ones). Practically, FedIRT-DP provides a single, auditable privacy-utility tool, enabling institutions to meet GDPR/FERPA-style requirements while maintaining competitive accuracy for federated calibration.

Beyond the federated learning and data privacy approach, other privacy-preserving techniques warrant future consideration. Cryptographic methods such as secure multi-party computation (SMPC) and homomorphic encryption enable joint analysis without revealing raw inputs \citep{damgaard2007efficient, rivest1978data}, yet their computational demands and incompatibility with discrete response formats limit practical implementation in IRT contexts. Traditional anonymization techniques (e.g., k-anonymity) risk distorting sparse categorical data when applied across distributed datasets \citep{sweeney2002k}. Although these alternatives offer formal privacy assurances, adapting them to psychometric models requires further methodological innovation.

There remain several future directions for improving FedIRT. First, the current application supports only unidimensional 2PL and PCM models; extending it to more complex IRT formulations such as the graded response model (GRM), nominal response model (NRM), and multidimensional frameworks would broaden its applicability. Additionally, integrating advanced machine learning methods, particularly deep neural networks, could improve estimation efficiency and model flexibility in realistic empirical settings when data generating process are not well defined which cause possible model misspecification or violations of distributional assumptions. Recent research has demonstrated promising results in applying deep learning algorithms to IRT models based on marginal maximum likelihood estimation \citep[e.g.,][]{urban2021deep,luo2024fitting,luo2025generative}.

From an implementation perspective, developing comprehensive documentation, detailed user guides, and intuitive interfaces for our methodological framework will be critical to supporting its broader adoption by educational institutions and policy agencies. Applying user-centered design principles and conducting extensive field testing with partner institutions will help ensure that the federated analysis tools we provide are not only technically sound but also accessible, reliable, and aligned with the real-world needs of diverse stakeholders. On this basis, although we have developed an initial version of a user-friendly Shiny application to facilitate ease of use, additional work is needed to further enhance its functionality, usability, and scalability. These continued efforts will be essential to ensure that the tool can effectively support decision-making and research across a range of educational settings.

Looking ahead, the FedIRT algorithm has the potential to serve a wide range of disciplines including education, psychology, social sciences, biomedical sciences, and beyond where distributed data analyses must reconcile rigorous statistical modeling with stringent privacy requirements. Its adaptability makes it a promising solution for any setting in which large-scale, distributed databases require joint analysis without compromising individual confidentiality with item response modelling. The DP extension, FedIRT-DP, strengthens this promise by coupling federated estimation with an auditable, user-level privacy guarantee and built-in robustness.

\bibliography{cite}
% \appendix
% \renewcommand{\theequation}{A\arabic{equation}}
% \setcounter{equation}{0}
% \renewcommand{\thesection}{\Alph{subsection}}
% \setcounter{section}{0}
% \section*{Appendix}\label{app:code}
\end{document}